%% file: main.tex
\documentclass{article}

     \PassOptionsToPackage{numbers, compress}{natbib}
\usepackage[main, final]{neurips_2026}

\usepackage[utf8]{inputenc} %
\usepackage[T1]{fontenc}    %
\usepackage[hidelinks]{hyperref}       %
\usepackage{url}            %
\usepackage{booktabs}       %
\usepackage{multirow}
\usepackage{amsfonts}       %
\usepackage{nicefrac}       %
\usepackage{microtype}      %
\usepackage{xcolor}         %
\usepackage{wrapfig}

\title{Sample Efficient Generative Molecular Optimization with Joint Self-Improvement}

\author{
\textbf{Serra Korkmaz}$^{1,2}$\thanks{Equal contribution.} \quad
\textbf{Adam Izdebski}$^{1,4,7}$\footnotemark[1] \quad
\textbf{Jonathan Pirnay}$^{3}$ \\
\textbf{Rasmus M{\o}ller-Larsen}$^{1,4,7}$ \quad
\textbf{Michal Kmicikiewicz}$^{1,4,7}$
\textbf{Pankhil Gawade}$^{1,4,7}$ \quad \\
\textbf{Maksim Pavlov}$^{1}$ \quad
\textbf{Dominik G. Grimm}$^{3,4}$ \quad
\textbf{Ewa Szczurek}$^{1,5,6,7}$ \quad
 \\
$^{1}$Institute of AI for Health, Computational Health Center, Helmholtz Munich\\
$^{2}$Technical University of Munich, TUM School of Natural Sciences \\
$^{3}$Technical University of Munich, TUM Campus Straubing for Biotechnology and Sustainability \\
$^{4}$Technical University of Munich, TUM School of Computation, Information and Technology \\
$^{5}$Faculty of Mathematics, Informatics and Mechanics, University of Warsaw \\
$^{6}$Faculty of Mathematics, Informatics and Statistics,  Ludwig-Maximilians-Universität München \\
$^{7}$ Munich Center for Machine Learning \\
\texttt{
\{serra.korkmaz, adam.izdebski, ewa.szczurek\}@helmholtz-munich.de}
}

\usepackage{adjustbox}

\usepackage{amsmath,amssymb,amsfonts,amsthm,mathtools}

\usepackage[nameinlink,noabbrev]{cleveref}

\usepackage{algorithm}
\usepackage{algpseudocode}

\newcommand{\x}{\mathbf{x}}
\newcommand{\X}{\mathcal{X}}

\newcommand{\refmodel}{p_\text{ref}(\x)}

\DeclareMathOperator*{\argmax}{arg\,max}

\newcommand{\E}{\mathbb{E}}

\newcommand{\R}{\mathbb{R}}

\newcommand{\logit}{\textrm{logit}\,}

\newcommand{\joint}{p_\theta(\x, y)}
\newcommand{\predictive}{p_\theta(y \mid \x)}
\newcommand{\generative}{p_\theta(\x)}
\newcommand{\reference}{p_{\mathrm{ref}}(\x)}

\theoremstyle{definition}

\begin{document}

\maketitle

\begin{abstract}
    Generative molecular optimization aims to design molecules with properties surpassing those of existing compounds. However, evaluating candidate molecules is expensive, so sample efficiency, i.e., generating optimized molecules within a fixed evaluation budget, is essential. Moreover, in the offline setting, where a surrogate model approximates candidate evaluation, sample inefficiency compounds with surrogate error, further limiting performance. To address these challenges, we introduce \textsc{Joint Self-Improvement} (\textsc{JSI}), a framework for sample-efficient online and offline molecular optimization. \textsc{JSI} combines \textbf{(i) a joint generative-predictive model} trained via a joint likelihood-based loss, which aligns molecule generation and surrogate modeling, and \textbf{(ii) inference-time self-improvement}, which iteratively biases the generative component of the joint model toward higher-scoring molecules, using direct objective function evaluations in the online setting and the predictive component in the offline setting. Experiments across docking score optimization in both online and offline settings, as well as antimicrobial peptide optimization, demonstrate that \textsc{JSI} outperforms state-of-the-art methods under limited evaluation budgets. Code is available at \url{https://github.com/szczurek-lab/jsi}.
\end{abstract}

\section{Introduction}
Generative molecular optimization (GMO) is a cornerstone of modern drug discovery, with the aim of generating molecules that improve upon existing compounds~\cite{fischer2019approaching,sridharan2022modern,caldas2025review, zhang2025artificial}. Yet, such molecules are exceedingly rare, and evaluating newly generated candidates remains prohibitively expensive, as it typically requires wet-lab experimentation~\cite{hughes2011principles, ozccelik2025generative}, making \emph{sample efficiency} essential~\cite{guo2024augmentedmemory}. 

Nevertheless, state-of-the-art GMO approaches typically rely on reinforcement learning (RL)~\cite{loeffler2024reinvent, guo2024augmentedmemory, guo2024saturnsampleefficientgenerativemolecular, shin2025offline}, in which a pre-trained chemical language model is treated as a policy and fine-tuned to generate high-scoring molecules. Since such molecules are exceedingly rare, sampling predominantly yields low-scoring candidates~\cite{izdebski2023novodrugdesignjoint}, rendering RL-based methods sample-inefficient~\cite{peters2006policy, zhao2011analysis, agarwal2020optimality, rengarajan2022reinforcement, vasan2024revisitingsparserewardsgoalreaching}.

Moreover, in settings representative of real-world drug discovery, where wet-lab experiments are required for candidate evaluation, it is often infeasible to evaluate molecules during optimization, an assumption made by \emph{online} molecular optimization~\cite{guo2024augmentedmemory, guo2024saturnsampleefficientgenerativemolecular}. In contrast, \emph{offline} molecular optimization assumes that candidates cannot be evaluated directly, relying instead on an offline dataset of pre-evaluated molecules to guide optimization~\cite{shin2025offline, kim2025offlinemodelbasedoptimizationcomprehensive}. The offline dataset is typically used to fit a \emph{surrogate}, which approximates candidate evaluation and serves as a plug-in replacement for the objective function in RL~\cite{shin2025offline}. However, estimating the objective beyond the offline dataset is inherently challenging, and optimizing against an imperfect surrogate risks \emph{overoptimization}, where the model exploits inaccuracies in the surrogate rather than improving the objective~\cite{levine2020offline, gao2023scaling, song2023hybrid, kim2025offlinemodelbasedoptimizationcomprehensive}.

\begin{figure*}[t!]
    \centering
    \includegraphics[width=\textwidth]{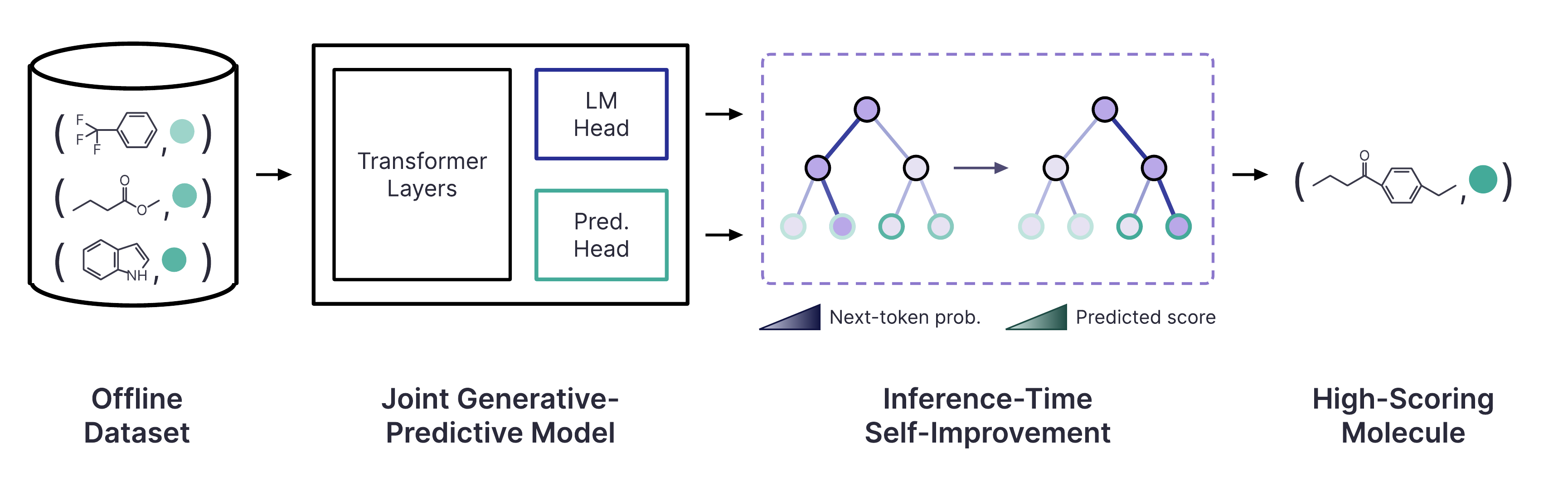}
    \vspace{-5ex}
    \caption{Overview of \textsc{JSI}. In the offline setting, a joint generative-predictive model, with a language modeling (LM) head for next-token probability prediction and a predictive head for property prediction, is fine-tuned on an offline dataset of molecules and their property scores, and the model self-improves by iteratively biasing next-token probabilities (blue) using predicted property scores (green), steering generation towards high-scoring molecules at inference time. In the online setting, ground-truth oracle evaluations are used in place of the predictive head. 
}
    \label{fig:overview}
    \vspace{-2ex}
\end{figure*}

To address these limitations, we introduce \textsc{Joint Self-Improvement} (\textsc{JSI}), a framework for sample-efficient online and offline GMO (Figure~\ref{fig:overview}). \textsc{JSI} aligns molecule generation and surrogate modeling within a \textbf{joint generative-predictive model} with shared parametrization, trained using the joint likelihood-based loss~\cite{izdebski2026synergistic}. At inference time, \textsc{JSI} generates optimized molecules by iteratively biasing the generative model's next-token probabilities towards high-scoring molecules, i.e., \textbf{inference-time self-improvement}~\cite{pirnay2024selfimprovement}. In the online setting, we generate molecules by biasing the generative model with ground-truth oracle evaluations and subsequently fine-tune the generative part of the model on the resulting samples. In the offline setting, where a surrogate replaces the ground-truth oracle, and only an offline dataset is available to guide optimization, we use the joint loss to fine-tune a pre-trained model and generate molecules by using the generative part of the joint model with the predictive part. We evaluate our framework on established benchmarks for sample-efficient molecular and antimicrobial peptide (AMP) optimization~\cite{guo2024saturnsampleefficientgenerativemolecular, shin2025offline, izdebski2026synergistic}. Our contributions are:
\begin{itemize}
\item We align molecule generation and surrogate modeling within a \textbf{joint generative-predictive model} with shared parametrization and trained using a joint likelihood-based loss.
\item We propose to sample from the joint model with \textbf{inference-time self-improvement}, iteratively biasing the generative component with oracle evaluations in the online setting, and with the predictive component in the offline setting.
\item We evaluate \textsc{JSI} on sample-efficient molecular optimization benchmarks, in both online and offline settings, as well as AMP optimization, achieving state-of-the-art results.
\end{itemize}

\section{Related Work}

GMO can be broadly divided into approaches that improve the generative model used to generate optimized molecules and those that improve the surrogate model used to guide optimization~\cite{kim2025offlinemodelbasedoptimizationcomprehensive}.

\paragraph{Generative Modeling for GMO} Early approaches to generative molecular optimization rely on genetic algorithms~\cite{jensen2019graph} or latent space optimization, where optimization is performed via gradient ascent or Bayesian optimization in a continuous latent representation of molecules~\cite{gomez2018automatic, jin2018junction, jin2020hierarchical, tripp2020sample, maus2022local, eckmann2022limo, lee2025nfbo}. More recently, RL methods are prevalent, including value-based approaches that optimize docking scores directly~\cite{jeon2020autonomous} and, in particular, policy gradient methods~\cite{shi2020graphaf, yang2021hit, jo2022score}. Following the growing emphasis on sample efficiency~\cite{gao2022sample}, subsequent work improves upon sample efficiency of RL by reusing off-policy data~\cite{guo2024augmentedmemory}, adopting Mamba-based backbones~\cite{guo2024saturnsampleefficientgenerativemolecular}, or combining RL with genetic algorithms~\cite{lee2024drug}. Similarly, \citet{kim2024genetic} incorporate domain-specific genetic operators into GFlowNets~\cite{bengio2023gflownet}. Beyond RL, \citet{lee2023exploring} perform conditional generation by leveraging gradients from a surrogate to guide a reverse diffusion process. Finally, \citet{pirnay2025graphxform} applies self-improvement, derived from round-wise provable policy improvement~\cite{pirnay2024selfimprovement}, to molecular graph construction in the online setting, and without evaluation in sample-efficient settings.

\paragraph{Surrogate Modeling for GMO}
A standard approach to surrogate modeling learns a predictor that approximates the objective function via mean squared error. Subsequent work improves upon this through adversarial training to obtain conservative predictions~\cite{trabucco2021coms}, invariant representation learning to enhance robustness under distribution shift~\cite{qi2022invariant}, bootstrapping strategies that iteratively augment the training set with self-generated samples~\cite{kim2023bootstrapped}, and learning-to-rank formulations that model relative orderings instead of absolute objective values~\cite{tan2024offline,shin2025offline}. 

\paragraph{Joint Modeling for GMO}
\citet{izdebski2026synergistic} jointly model molecule generation and property prediction, showing that shared parametrization aligns the generative and predictive components, enabling property-conditioned molecule generation. However, they rely on rejection sampling to draw high-quality candidates, limiting exploration to high-density regions of the generative model. \citet{pirnay2025graphxform} apply self-improvement to molecular graph construction, but act solely on the generative model without surrogate integration. \textsc{JSI} combines both paradigms by leveraging the predictive component of the joint model to directly guide generation via self-improvement, enabling efficient sampling beyond the generative model's support in both online and offline settings.

\section{Background}

\paragraph{Molecular Optimization} The goal of \emph{molecular optimization} is to propose a \emph{molecule} ${\x}^* \in \X$ that maximizes an \emph{objective function} (\emph{oracle}) $f: \X \to [0, f_{\max}]^M$, where $f_{\max} \in \R_+$. When $M=1$, molecular optimization reduces to a single-objective optimization problem 
\begin{equation}
    {\x}^* = \argmax_{\x \in \X} f(\x).
\end{equation}
In practice, the chemical space $\X$ is further restricted by feasibility criteria, such as synthetic accessibility or drug-likeness. As evaluating the objective $f$ is costly, e.g., a wet-lab assay or docking simulation, molecular optimization prioritizes \emph{sample efficiency}, i.e., minimizing the number of objective function evaluations (oracle calls) needed to propose an optimized molecule $\x^*$.

Molecular optimization additionally depends on whether the objective $f$ can be evaluated during optimization. In the \emph{online} setting, the optimization process iteratively queries $f$ and adapts based on the feedback. However, as $f$ is costly to evaluate, the online setting is often infeasible in practice. In contrast, the \emph{offline} setting~\cite{kim2025offlinemodelbasedoptimizationcomprehensive} assumes access only to an \emph{offline dataset} $\mathcal{D}_{\mathrm{offline}} = \{(\x^{(n)}, f(\x^{(n)}))\}_{n=1}^{N}$.

\paragraph{Reinforcement Learning-based Molecular Optimization} State-of-the-art molecular optimization methods~\cite{guo2024augmentedmemory, guo2024saturnsampleefficientgenerativemolecular, shin2025offline} predominantly rely on a variant of reinforcement learning, in which a generative model $\generative$ is fine-tuned via the \emph{KL-regularized objective}
\begin{equation}\label{eq:kl-regularized-objective}
    J_{f, \beta}(\theta) = \mathbb{E}_{p_\theta(\x)}[f(\x)]
    - \beta\,\mathrm{KL}\!\left(p_\theta(\x)\,\|\,\reference\right),
\end{equation}
where $\beta > 0$ controls the drift from a \emph{reference model} $\reference$, typically a pre-trained chemical language model. More generally, this is an instance of \emph{reinforcement learning fine-tuning} (RLFT)~\cite{murphy2025reinforcementlearningoverview}. 

Despite widespread adoption, optimizing the KL-regularized objective $J_{f, \beta}$ is inherently sample-inefficient. In regimes typical of molecular optimization, molecules with high values of the objective $f$ are rare and lie in low-density regions of $\reference$. In consequence, unconditional sampling $\x \sim p_\theta(\x)$ predominantly yields molecules with low values of $f$ when $\generative$ is close to $\reference$~\cite{izdebski2023novodrugdesignjoint}. Yet, the Monte Carlo gradient estimator~\cite{tang2025pitfallskldivergencegradient}
\begin{equation}\label{eq:mc-gradient}
\nabla_{\theta} J_{f, \beta}(\theta) \approx \frac{1}{|B|} \sum_{\x \in B} \left(f(\x) - \beta\log\frac{p_\theta(\x)}{\reference}\right) \nabla_\theta \log p_\theta(\x),
\end{equation}
where $B = \{{\x}^{(b)}\}_{b=1}^{|B|} \sim p_\theta(\x)$, requires the batch $B$ to contain molecules with high values of $f$ to provide a meaningful training signal, rendering fine-tuning sample-inefficient, especially early in training~\cite{peters2006policy, zhao2011analysis, agarwal2020optimality, song2023hybrid, vasan2024revisitingsparserewardsgoalreaching}. In particular, in Appendix~\ref{appendix:REINVENT-gradient}, we show that REINVENT-style algorithms~\cite{loeffler2024reinvent} inherit these limitations.

\paragraph{Surrogate Modeling} In the offline setting, where the objective $f$ cannot be evaluated, a surrogate $\hat{f}: \X \to [0, f_{\max}]$ is trained on the offline dataset to approximate $f$ and used as a plug-in replacement for the objective~\cite{kim2025offlinemodelbasedoptimizationcomprehensive}. When combined with the KL-regularized objective, this amounts to optimizing $J_{\hat{f}, \beta}$ in place of $J_{f, \beta}$ (Eq.~\eqref{eq:kl-regularized-objective}). However, as the surrogate $\hat{f}$ is an imperfect proxy for $f$, particularly outside the support of $\mathcal{D}_{\mathrm{offline}}$, optimizing $J_{\hat{f}, \beta}$ for too long, e.g., due to sample inefficiency of the underlying optimization algorithm, may cause the expected true objective $\E_{\generative}[f(\x)]$ to deteriorate even as $\E_{\generative}[\hat{f}(\x)]$ continues to improve, a phenomenon known as \emph{overoptimization}~\cite{gao2023scaling}.

\section{Joint Self-Improvement}

We introduce \textsc{JSI}, a unified framework for sample-efficient molecular optimization. First, we formulate molecular optimization as a conditional sampling problem. Next, we define a joint model $p_\theta(\x, y)$ that unifies the generative model $p_\theta(\x)$ and the surrogate $p_\theta(y \mid \x)$~within a shared parametrization $\theta$. Finally, we propose to sample from the joint model $\joint$ by self-improving the joint model at inference time.

\paragraph{Molecular Optimization as Conditional Sampling}
Molecular optimization can be formulated as a conditional sampling problem. Indeed, maximizing the KL-regularized objective $J_{f, \beta}(\theta)$ in Eq.~\eqref{eq:kl-regularized-objective} admits a closed-form solution~\cite{knoblauch2019generalizedvariationalinferencearguments,korbak2022reinforcementlearningdistributionmatching,rafailov2023dpo, murphy2025reinforcementlearningoverview}, given by 
\begin{equation}\label{eq:gibbs-posterior}
    p^*(\x; f, \beta) = Z_{f,\beta}^{-1}\, \reference\exp\!\left(\beta^{-1} f(\x)\right)
\end{equation}
where $Z_{f, \beta} = \E_{\refmodel}[\exp\!\left(\beta^{-1} f(\x)\right)]$ is the normalizing constant and $\beta>0$ controls the drift from the reference model $\reference$. However, evaluating the objective $f$ is expensive, and directly sampling from $p^*(\x; f, \beta)$ is infeasible, as the constant $Z_{f, \beta}$ is intractable.

\paragraph{Joint Generative-Predictive Model} To approximate the objective $f$ with a surrogate, we define a \emph{joint generative-predictive model}
\cite{nalisnick2019hybrid, grathwohl2020classifier, izdebski2026synergistic}
\begin{equation}\label{eq:joint-model}
   p_\theta(\x, y) = p_\theta(y \mid \x)\, p_\theta(\x),
\end{equation}
where $p_\theta(y \mid \x)$ is the surrogate and $p_\theta(\x)=\prod_{t=1}^Tp_\theta(x_t \mid {\x}_{1:t-1})$ is an \emph{autoregressive} generative model. We train the joint model via \emph{joint loss}, i.e., the joint log-likelihood
\begin{equation}\label{eq:joint-loss}
   \mathcal{L}_{\lambda}(\theta)
   = - \sum_{(\x, y) \in \mathcal{D}}
   \Bigl( (1-\lambda)\log p_\theta(\x) + \lambda \log p_\theta(y \mid \x) \Bigr),
\end{equation}
where $\lambda \in [0, 1]$ controls the relative strength of the generative and predictive counterparts. Note that when $\lambda=0$ or $\lambda=1$, the joint loss reduces to generative-only or predictive-only updates, respectively.

The advantages of training a joint model are two-fold. As opposed to optimizing the KL-regularized objective $J(\theta)$ in Eq.~\eqref{eq:kl-regularized-objective}, the joint loss yields stable likelihood-based updates, as the gradients of $\mathcal{L}_{\lambda}$ do not depend on sampling optimized examples~\cite{izdebski2026synergistic}. Additionally, in the offline setting, both the predictive and the generative components are trained jointly on the same offline dataset $\mathcal{D}_{\mathrm{offline}}$, and share parameters $\theta$, effectively aligning the generative distribution with the surrogate~\cite{izdebski2026synergistic}.

\paragraph{Inference-Time Self-Improvement} To sample optimized molecules $\x^*$ from the joint model $\joint$ at \emph{inference-time}, we propose to \emph{self-improve} the generative component $\generative$ of the joint model $\joint$, by perturbing its next-token probabilities with the predictive component~$\predictive$, or directly with $f(\x)$ in the online setting. 

To achieve this, we define the \emph{individual advantage} \cite{pirnay2024selfimprovement} of token \(x_t \in \mathcal{V}\), given a prefix \(\x_{1:t-1}\), as
\begin{equation}\label{eq:expected-advantage}
        \textrm{Adv}_\theta(x_t \mid {\x}_{1:t-1}) := \E_{p_\theta({\x}_{t+1:T} \mid x_t, {\x}_{1:t-1})}\big[p_\theta(y \mid \x)\big] - \E_{p_\theta({\x}_{t:T}\mid {\x}_{1:t-1})}\big[p_\theta(y \mid \x)\big],
\end{equation}
where both expectations are taken over completed sequences $\x=(x_1, \ldots, x_T)$, estimated via the generative part \(p_\theta(\x)\). Using the individual advantage, we define the logits of the \emph{perturbed model} $p_{\theta'}(\x)$, given by locally tilting the next-token log-probabilities of the generative model $p_{\theta}(\x)$
\begin{equation}
\label{eq:perturbed-logits}
\text{logit}\, p_{\theta'}(x_t \mid {\x}_{1:t-1})
:= 
\log p_\theta(x_t \mid {\x}_{1:t-1}) \\
+
\sigma\, \textrm{Adv}_\theta(x_t \mid {\x}_{1:t-1}),
\end{equation}
where $\sigma > 0$ is the \emph{step size}. Depending on the step size $\sigma$, the perturbed model $p_{\theta'}(\x)$ interpolates between generating molecules that are likely under the generative model $p_\theta(\x)$, similar to methods based on rejection-sampling~\cite{izdebski2026synergistic}, and exploring low-density regions of $p_\theta(\x)$. Yet, estimating individual advantages can result in high-variance updates, and estimating advantages over completed sequences%
has been shown to provide stronger empirical performance~\cite{pirnay2024selfimprovement}. Consequently, the practical implementation of our method relies on global advantages, as detailed in the next section.

\begin{figure*}[t]
    \centering
    \includegraphics[width=\linewidth]{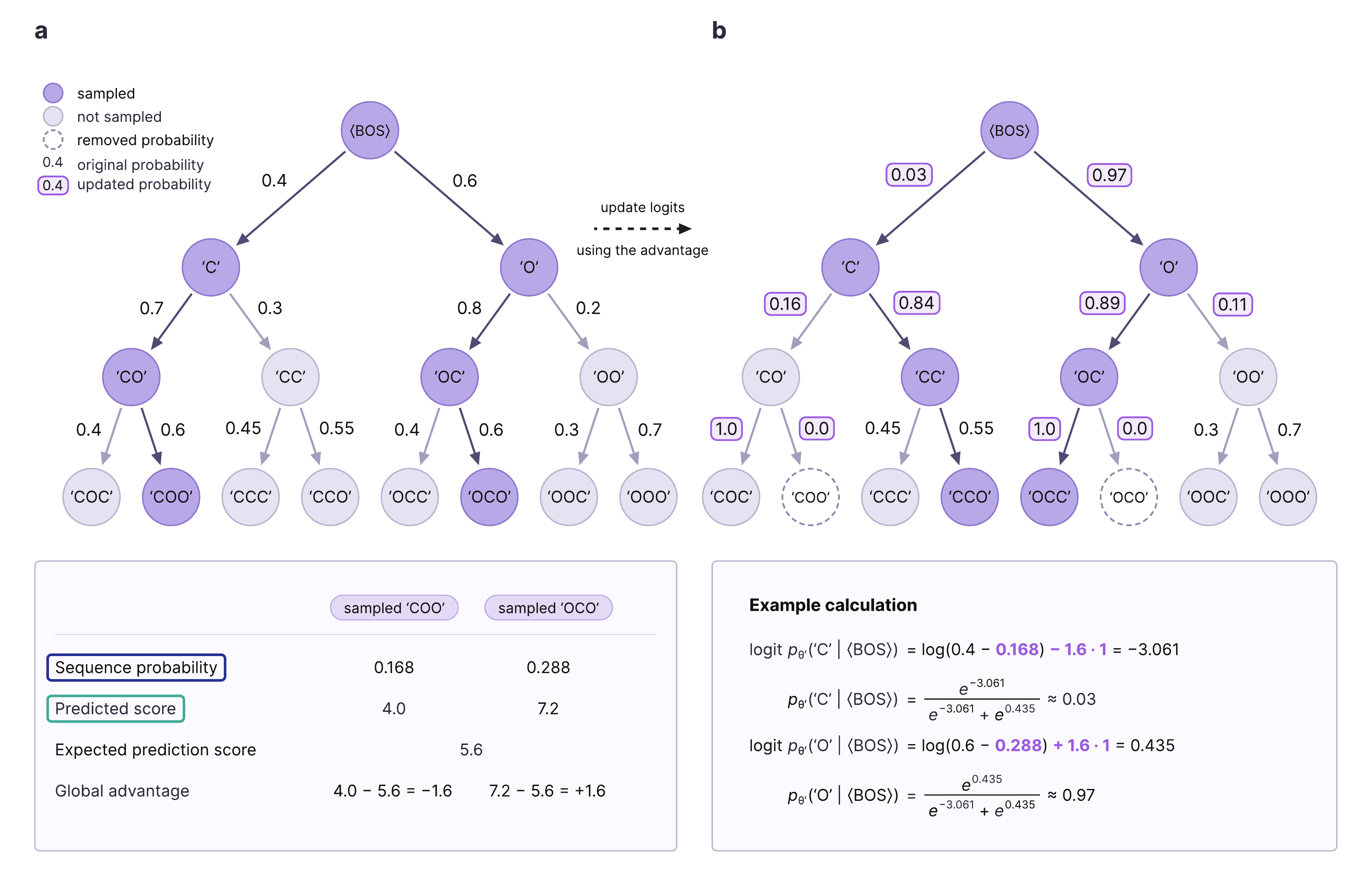}
    \vspace{-6ex}
    \caption{
    Example logit update in Eq.~\eqref{eq:perturbed-model-practical} with $K=2$, $\sigma=1$ and $\mathcal{V} = \{C, O\}$. \textsc{JSI} uses SBS to autoregressively generate $K$ molecules, inducing a tree where each node is a partial sequence and edges carry next-token probabilities. The predictive head scores each completed sequence, and the scores are aggregated into a global advantage that biases next-token probabilities. Panel \textbf{(a)} shows the initial probabilities, while panel \textbf{(b)} shows probabilities biased towards high-scoring branches.
    }
    \vspace{-2.5ex}
    \label{figure:sampling}
\end{figure*}

\paragraph{Practical Implementation}
In practice, to sample from the joint model $\joint$ (Figure~\ref{figure:sampling}), we follow~\cite{pirnay2024selfimprovement} and first draw $\{{\x}^{(k)}\}_{k=1}^K \sim p_{\theta}(\x)$ using Stochastic Beam Search (SBS), without replacement. By implementing SBS using the \emph{Gumbel-Top-k trick}, sampling requires only a number of model evaluations that scales linearly with the number of samples $K$ and sequence length $T$~\cite{kool2019stochastic}. 
Next, for every sequence ${\x}^{(k)} \in \{{\x}^{(k)}\}_{k=1}^K$, we compute its \textit{global advantage}
\begin{equation}\label{eq:definition-global-advantage}
    \textrm{Adv}_\theta({\x}^{(k)}) = p_\theta(y \mid {\x}^{(k)}) - \mu({\x}^{(1)}, \ldots, {\x}^{(K)}),
\end{equation}
where $\mu({\x}^{(1)}, \ldots, {\x}^{(K)})$ is a consistent estimator of the \emph{expected prediction score} $\E_{p_\theta(\x)}[p_\theta(y \mid \x)]$ \cite{hesterberg1988advances, kool2019stochastic, pirnay2024selfimprovement}. Define $S({\x}_{1:t}^{(k)})$ to be the set of all sequences $\x \in \{{\x}^{(k)}\}_{k=1}^K$ that share the prefix ${\x}_{1:t}^{(k)}$. Finally, we define the logits, i.e., unnormalized log-probabilities, of the perturbed generative component $p_{\theta'}(\x)$, for each ${\x}^{(k)} \in \{{\x}^{(k)}\}_{k=1}^K$ and $t \in [T]$, as
\begin{equation}\label{eq:perturbed-model-practical}
        \logit p_{\theta'}
        \bigl(x^{(k)}_t \mid {\x}^{(k)}_{1:t-1}\bigr) :=
        \log\Bigl(p_\theta(x^{(k)}_t
        \mid {\x}^{(k)}_{1:t-1})-\hspace{-2.5ex}\sum_{\x \in S({\x}^{(k)}_{1:t})}\hspace{-2ex}
        p_\theta({\x} \mid {\x}_{1:t-1}^{(k)})\Bigr) + \sigma\hspace{-2ex}\sum_{\x \in S({\x}^{(k)}_{1:t})}\hspace{-2ex} \textrm{Adv}_{\theta}({\x}),
\end{equation}
where the logarithm is well-defined as the total probability mass of sequences in $S(\x_{1:t}^{(k)})$ is upper-bounded by the conditional probability $p_\theta(x_t^{(k)} \mid \x_{1:t-1}^{(k)})$ (see Appendix~\ref{appendix:perturbed-logits} for a derivation and Appendix~\ref{appendix:numerical-stability} for numerical stability). Intuitively, $\sum_{\x \in S({\x}^{(k)}_{1:t})}p_\theta({\x} \mid {\x}_{1:t-1}^{(k)})$ ensures that we sample without replacement~\cite{kool2019stochastic}, while $\sum_{\x \in S({\x}^{(k)}_{1:t})}\textrm{Adv}_{\theta}({\x})$ biases the generative component towards samples with a high prediction score, according to the predictive component of the joint model. We iteratively repeat the above procedure for a fixed number of iterations, and summarize the complete procedure in Algorithm~\ref{alg:joint-self-improvement} (Appendix~\ref{appendix:sampling-with-jsi}).

\section{Experiments}\label{section:experiments}

To evaluate \textsc{JSI} in sample-efficient molecular optimization, we benchmark our method on docking score optimization in both online (Section~\ref{section:online-molecular-optimization}) and offline (Section~\ref{section:offline-molecular-optimization}) settings. To demonstrate the generalizability of our method beyond docking score optimization, we design an AMP optimization task, in which we consider offline datasets with a restricted proportion of high-scoring peptides, i.e., peptides with MIC values below a predefined threshold, further increasing task difficulty (Section~\ref{section:amp}). Additionally, we quantify the individual contributions of joint modeling and inference-time self-improvement through ablation studies (Section~\ref{section:ablations}). Further experiments, covering surrogate accuracy, hyperparameter sensitivity, reduced evaluation budgets and offline dataset sizes, and sampling cost, are reported in Appendix~\ref{appendix:additional-experiments}, and experimental details in Appendix~\ref{appendix:experimental-details}.

\subsection{Docking Score Optimization}

\paragraph{Experimental Setup} We evaluate molecular optimization on docking score optimization across five protein targets: PARP1, FA7, JAK2, BRAF, and 5HT1B, while imposing feasibility constraints on drug-likeness~\cite{bickerton2012quantifying} (QED) and synthetic accessibility~\cite{ertlSilicoGenerationNovel2018} (SA), under an oracle budget of 3{,}000 evaluations~\cite{lee2024drug, guo2024saturnsampleefficientgenerativemolecular}, a more stringent regime than the 10{,}000 evaluations used by~\citet{gao2022sample}.

\paragraph{Metrics} Following \cite{lee2024drug, guo2024saturnsampleefficientgenerativemolecular}, we measure \textbf{Hit Ratio} (\%), defined as the fraction of sampled hits, i.e., molecules with a docking score strictly better than the median of known actives~\cite{guo2024saturnsampleefficientgenerativemolecular}, QED > 0.5, and SA < 5. Additionally, we evaluate internal diversity (\textbf{IntDiv1})~\cite{benhenda2017chemgan} and oracle burden (\textbf{OB}), where OB($n$) denotes the number of oracle calls required to obtain $n$ unique hits. For each metric, we report the mean and standard deviation calculated across 10 random seeds.

\begin{table*}[t!]
    \caption{Molecular optimization performance on docking score optimization in the online setting. We mark the best method \textbf{bold}.}
    \begin{center}
    \begin{scriptsize}
    \begin{sc}
    \begin{tabular}{@{}lccccc@{}}
    \toprule
    & \multicolumn{5}{c}{Target Protein, Hit Ratio ($\uparrow$)} \\
    \cmidrule{2-6}
    Method       & parp1 & fa7 & 5ht1b & braf & jak2 \\
    \midrule
    ZINC250K
    & 3.993 (0.355) & 1.097 (0.192) & 24.26 (0.622) & 1.020 (0.193) & 6.183 (0.344) \\
    REINVENT
    & 4.693 (1.776) & 1.967 (0.661) & 26.047 (2.497) & 2.207 (0.800) & 5.667 (1.067) \\
    JT-VAE
    & 3.200 (0.348) & 0.933 (0.152) & 18.044 (0.747) & 0.644 (0.157) & 5.856 (0.204) \\
    GraphAF
     & 0.822 (0.113) & 0.011 (0.016) & 6.978 (0.952) & 1.422 (0.556) & 1.233 (0.284) \\
     MORLD
     & 0.047 (0.050) & 0.007 (0.013) & 0.893 (0.758) & 0.047 (0.040) & 0.227 (0.118) \\
     HierVAE
     & 1.180 (0.182) & 0.033 (0.030) & 0.740 (0.371) & 0.367 (0.187) & 0.487 (0.183) \\
    FREED-QS
    & 5.960 (0.902) & 1.687 (0.177) & 23.140 (2.422) & 3.880 (0.623) & 7.653 (1.373) \\
     LIMO
     & 0.456 (0.057) & 0.044 (0.016) & 1.200 (0.178) & 0.278 (0.134) & 0.711 (0.329) \\
     GDSS
     & 2.367 (0.316) & 0.467  (0.112) & 6.267 (0.287) & 0.300 (0.198) & 1.367 (0.258) \\
    MOOD
    & 7.260 (0.764) & 0.787 (0.128) & 21.427 (0.502) & 5.913 (0.311) & 10.367 (0.616) \\
    Augmented Memory
    & 16.966 (3.224) & 2.637 (0.860) & 52.016 (2.302) & 8.307 (1.714) & 21.548 (4.938) \\
    GEAM
    & 45.158 (2.408) & 20.552 (2.357) & 47.664 (1.198) & 30.444 (1.610) & 46.129 (2.073) \\
    SATURN & 57.981 (18.537) & 14.527 (9.961) & 68.185 (3.400) & 38.999 (10.114) & 60.827 (11.502) \\
    NF-BO
    & 18.641 (8.072)&	5.305 (3.568)&	53.185 (3.606)&	16.593 (3.577)&	29.537 (6.949) \\

    \textsc{JSI (ours)} & \textbf{67.280 (7.961) } & \textbf{36.587 (10.270)} & \textbf{91.357 (3.204)} & \textbf{58.547 (11.456)} & \textbf{71.863 (6.127)} \\
    \bottomrule
  \end{tabular}
  \end{sc}
  \end{scriptsize}
  \end{center}
  \label{experiments:table:hit-ratio-online}
  \vspace{-4ex}
\end{table*}

\begin{figure*}[b!]
    \vspace{-2ex}
    \centering  \includegraphics[width=\textwidth]{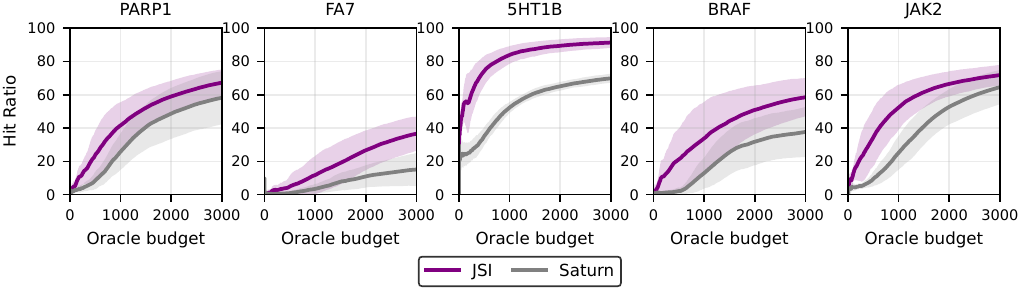}
    \vspace{-3ex}
    \caption{Hit ratio versus oracle budget across five protein targets. Curves show the mean across 10 seeds for \textsc{JSI} (purple) and SATURN (gray); shaded regions indicate $\pm 1$ standard deviation over~seeds.}
    \label{fig:hit-ratio-oracle-budget-online}
\end{figure*}

\subsubsection{Sample-Efficient Online Molecular Optimization}\label{section:online-molecular-optimization}
To evaluate the sample efficiency of \textsc{JSI} in the online setting, we follow the experimental setup of~\citet{guo2024saturnsampleefficientgenerativemolecular}. We generate optimized molecules, by using Hyformer~\cite{izdebski2026synergistic} pre-trained on ZINC250K~\cite{sterling2015zinc}, as the backbone for \textsc{JSI}, and defining the reward $R(\x)$ to calculate the global advantage of a sequence $\x$, in place of the prediction score in Eq.~\eqref{eq:definition-global-advantage}, as
\begin{equation}\label{eq:reward-online}
    R(\x) = \frac{1}{3}\Bigl(-\mathrm{DS}^{\textrm{z}}(\x) - \mathrm{SA}^{\textrm{z}}(\x) + \mathrm{QED}^{\textrm{z}}(\x)\Bigr),
\end{equation}
where $\mathrm{DS}^{\textrm{z}}$, $\mathrm{SA}^{\textrm{z}}$ and $\mathrm{QED}^{\textrm{z}}$ are Z-score normalized \cite{hu2025reinforce++} values of DS, SA and QED, respectively, calculated using the ground-truth oracle. Next, we use the generated molecules to fine-tune the backbone using the generative-only part of the joint loss $\mathcal{L}_{\lambda = 0}$ (Eq.~\eqref{eq:joint-loss}). We repeat the above procedure until the oracle budget of 3{,}000 evaluations is exhausted. Further experimental details, together with an overview of the different optimization settings (Figure~\ref{fig:overview-experimental-setting}), are in Appendix~\ref{appendix:online-optimization}. We analyze the choice of reward normalization and aggregation in Appendix~\ref{appendix:reward-analysis}, and the relative weighting of the individual properties in Appendix~\ref{appendix:weighted-reward}.

We compare \textsc{JSI} with the baselines reported by~\citet{guo2024saturnsampleefficientgenerativemolecular}: ZINC250K~\cite{sterling2015zinc}, molecules sampled at random from the pre-training dataset; REINVENT~\cite{loeffler2024reinvent}, a SMILES-based generative model fine-tuned with reinforcement learning towards an augmented likelihood; JT-VAE~\cite{jin2018junction}, a graph-based variational autoencoder; GraphAF~\cite{shi2020graphaf}, a flow-based autoregressive graph generator; MORLD~\cite{jeon2020autonomous}, which optimizes molecules with deep Q-learning on docking rewards; HierVAE~\cite{jin2020hierarchical}, a hierarchical motif-by-motif decoder; FREED-QS~\cite{yang2021hit}, a fragment-based reinforcement learning approach with experience replay; LIMO~\cite{eckmann2022limo}, which optimizes a VAE latent space via a differentiable property predictor; GDSS~\cite{jo2022score}, a score-based graph diffusion model; MOOD~\cite{lee2023exploring}, a conditional diffusion-based sampling method; Augmented Memory~\cite{guo2024augmentedmemory}, an extension of REINVENT that combines experience replay with SMILES augmentation; GEAM~\cite{lee2024drug}, which integrates reinforcement learning with a genetic algorithm for goal-aware fragment-based optimization; and SATURN~\cite{guo2024saturnsampleefficientgenerativemolecular}, a REINVENT-style optimization framework with a Mamba backbone. We additionally evaluate NF-BO~\cite{lee2025nfbo}, a latent Bayesian optimization method built on an autoregressive normalizing flow, which we run ourselves.

\textsc{JSI} consistently outperforms all baselines across all protein targets (Table~\ref{experiments:table:hit-ratio-online}). REINVENT achieves a low overall Hit Ratio, indicating limited ability to leverage oracle feedback under strict evaluation budgets. In contrast, methods designed to improve sample efficiency, such as Augmented Memory and SATURN, yield substantial gains, but still underperform compared to \textsc{JSI}. Investigation of performance across all oracle budgets up to 3{,}000 (Figure~\ref{fig:hit-ratio-oracle-budget-online}) demonstrates that \textsc{JSI} improves over state-of-the-art SATURN at any given budget. This suggests that \textsc{JSI} provides a strong inductive bias for sample-efficient online molecular optimization, rapidly identifying hit molecules, especially at the early stages of optimization.
Finally, \textsc{JSI} consistently reduces oracle burden across all protein targets, at the cost of slightly decreased diversity, with the differences being particularly pronounced for harder targets such as FA7 and BRAF (Table~\ref{tab:oracle_burden_online}). We additionally report the validity, novelty and similarity to the training set of the generated molecules in Appendix~\ref{appendix:online-additional-metrics}.

\begin{table*}[t]
    \caption{Oracle burden in the online setting. We mark the best method \textbf{bold}.}
    \begin{center}
    \begin{scriptsize}
    \begin{sc}
    \begin{tabular}{@{}llccccc@{}}
    \toprule
    & & \multicolumn{5}{c}{Target Protein} \\
    \cmidrule{3-7}
    &Method & parp1 & fa7 & 5ht1b & braf & jak2 \\
    \midrule

     \multirow{1}{*}{SATURN}
    &OB(1) ($\downarrow$)   & 35 (28)  & 201 (172) & 4 (3) & 130 (144) & 30 (29) \\
    &OB(10) ($\downarrow$)  & 282 (116) & 715 (187) & 45 (13) & 517 (167) & 227 (93) \\
    &OB(100) ($\downarrow$) & 721 (137) & 1408 (304) & 347 (49) & 1047 (301) & 720 (131) \\
    & IntDiv1 ($\uparrow$) 
    & \textbf{0.627 (0.023)} 
    & \textbf{0.597 (0.056)} 
    & \textbf{0.665 (0.032)} 
    & \textbf{0.572 (0.063)} 
    & \textbf{0.641 (0.056)} \\

    \midrule

   \multirow{1}{*}{\textsc{JSI (ours)}}
    &OB(1) ($\downarrow$)   & \textbf{25 (21)} & \textbf{106 (77)} & \textbf{3 (2)} & \textbf{68 (72)} & \textbf{13 (9)} \\
    & OB(10) ($\downarrow$)  
    & \textbf{186 (143)} 
    & \textbf{424 (217)} 
    & \textbf{25 (6)} 
    & \textbf{210 (134)} 
    & \textbf{94 (29)} \\

    & OB(100) ($\downarrow$) 
    & \textbf{513 (171)} 
    & \textbf{1106 (388)} 
    & \textbf{184 (47)} 
    & \textbf{632 (330)} 
    & \textbf{401 (139)} \\
    &IntDiv1 ($\uparrow$) & 0.571 (0.037) & 0.571 (0.019) & 0.595 (0.053) & 0.554 (0.038) & 0.586 (0.038) \\

    \bottomrule
    \end{tabular}
    \end{sc}
    \end{scriptsize}
    \end{center}
    \label{tab:oracle_burden_online}
    \vspace{-2ex}
\end{table*}

\subsubsection{Sample-Efficient Offline Molecular Optimization}\label{section:offline-molecular-optimization}

To evaluate the sample efficiency of \textsc{JSI} in the offline setting, we follow the experimental setup of~\citet{shin2025offline}. For each of $5$ protein targets and $10$ random seeds, we construct an offline dataset by randomly selecting 1{,}500 molecules from ZINC250K and computing their docking score, SA and QED. Next, we fine-tune Hyformer, pre-trained on ZINC250K, using the joint loss $\mathcal{L}_{\lambda = 0.25}$ (Eq.~\eqref{eq:joint-loss}). In contrast to the online setting, where ground-truth oracle evaluations are available, here we update both the generative and predictive parts of the joint model. To generate optimized molecules, we iteratively bias the generative part of the joint model using the reward
\begin{equation}\label{eq:reward-offline}
    \hat{R}(\x) = \frac{1}{3}\Bigl(-\hat{\mathrm{DS}}^{\textrm{z}}(\x) - \hat{\mathrm{SA}}^{\textrm{z}}(\x) + \hat{\mathrm{QED}}^{\textrm{z}}(\x)\Bigr),
\end{equation}
where $\hat{\mathrm{DS}}^{\textrm{z}}$, $\hat{\mathrm{SA}}^{\textrm{z}}$ and $\hat{\mathrm{QED}}^{\textrm{z}}$ are the Z-score normalized docking score, SA and QED, respectively, estimated by the predictive part of the joint model. Finally, we use the remaining oracle budget of 1{,}500 to evaluate the generated molecules. Further experimental details are in Appendix~\ref{appendix:offline-optimization}.

Following~\cite{shin2025offline}, we compare \textsc{JSI} against: GraphGA~\cite{jensen2019graph}, a graph-based genetic algorithm; REINVENT~\cite{loeffler2024reinvent}; SATURN~\cite{guo2024saturnsampleefficientgenerativemolecular}; GeneticGFN~\cite{kim2024genetic}, combining genetic algorithms with GFlowNets; and RaM~\cite{tan2024offline}, a ranking-based offline optimization method that replaces regression-based surrogate learning with learning-to-rank objectives. We do not compare against MolStitch~\cite{shin2025offline} due to reproducibility limitations.

As in the online setting, here again \textsc{JSI} outperforms all baselines across all protein targets (Table~\ref{table:offline-optimization}). REINVENT and SATURN yield consistently low Hit Ratios across all targets, highlighting their sample inefficiency. RaM represents a strong baseline, substantially improving upon earlier methods. Additionally, to qualitatively validate the performance of \textsc{JSI} in the offline setting, we show binding poses of the best hit molecules, i.e., with the lowest docking scores, in Figure~\ref{fig:docked_poses}. 

\begin{table*}[b]
\centering
\vspace{-4ex}
\caption{Molecular optimization performance on docking score optimization in the offline setting. We mark the best method \textbf{bold}.}
\vspace{1ex}
\label{table:offline-optimization}
\begin{sc}
\begin{scriptsize}
\begin{tabular}{lccccc}
\toprule
& \multicolumn{5}{c}{Target Protein, Hit Ratio ($\uparrow$)} \\
\cmidrule{2-6}
Method & parp1 & fa7 & 5ht1b & braf & jak2 \\
\midrule
GraphGA & 29.631 (20.478)	& 0.913 (0.942)	& 73.875 (4.945)	& 3.340 (4.056)	& 28.912 (14.910) \\
REINVENT & 2.686 (0.612) & 0.813 (0.252)  & 19.653 (0.937) & 0.693 (0.204) & 4.020 (0.616) \\
SATURN & 6.319 (0.845) & 0.733 (0.270) & 32.287 (1.687) & 1.927 (0.353) & 7.667 (1.279) \\
GeneticGFN        & 8.893 (1.532) & 0.767 (0.252) & 56.910 (2.730) & 2.760 (0.907) & 10.693 (1.927) \\
RaM               & 38.373 (5.196) & 5.712 (5.382) & 62.932 (9.472) & 10.732 (4.771) & 33.189 (7.850) \\
\textsc{JSI (ours)} & \textbf{45.628 (11.653)} & \textbf{5.831 (2.040)} & \textbf{76.940 (4.442)} & \textbf{21.637 (4.607)} & \textbf{42.001 (4.344)} \\
\bottomrule
\end{tabular}
\end{scriptsize}
\end{sc}
\end{table*}

\begin{figure*}[t]
    \centering
    \includegraphics[width=\textwidth]{Figure4_v2.png}
    \vspace{-3ex}
    \caption{Binding poses of best molecules generated by \textsc{JSI} in the offline optimization setting. 
}
    \label{fig:docked_poses}
    \vspace{-2ex}
\end{figure*}

\subsection{AMP Optimization}\label{section:amp}

\paragraph{Experimental Setup} We evaluate \textsc{JSI} on the AMP optimization task, adapting the setup of~\citet{izdebski2026synergistic} to the sample-efficient setting by restricting the evaluation budget to 1K peptide sequences. In particular, we use Hyformer~\cite{izdebski2026synergistic} as the backbone for \textsc{JSI}, pre-trained on 3.5M peptide sequences and subsequently on 1M AMP sequences. The offline dataset consists of 3{,}444 peptides with associated MIC values against \textit{E.\ coli}, following~\citet{Szymczak2023HydrAMP}. While prior work defines activity labels by thresholding MIC values, labeling peptides with MIC $\leq 10^{1.5} \simeq 32\,\mu\mathrm{g}/\mathrm{mL}$ as hits, we treat MIC prediction as a regression task and use the continuous values during training and optimization. The thresholded labels are used only to characterize the dataset composition. To further increase task difficulty, we construct imbalanced offline datasets that retain the 1{,}318 non-hit peptides but restrict the proportion of hits to 1\% (13 peptides) and 0\% (no hits). For each method, we report the mean and 95\% confidence intervals estimated via bootstrapping.

\paragraph{Metrics}
For evaluation, we consider {\textbf{perplexity}}~\cite{torres2025ampdiffusion}, {\textbf{diversity}}, and {\textbf{fitness}}~\cite{li2024foundationamp}, along with three state-of-the-art \textit{in silico} oracles: \textbf{HydrAMP$_{\text{MIC}}$}~\cite{Szymczak2023HydrAMP}, which predicts the probability that a peptide is active against \textit{E.\ coli}, and \textbf{AMPlify}~\cite{li2022amplify} and \textbf{AmPEPpy}~\cite{lawrence2020ampeppy} for predicting the probability that a peptide is antimicrobial.

\subsubsection{Sample-Efficient Offline AMP Optimization}\label{section:offline-amp-optimization}

To sample optimized AMPs with \textsc{JSI}, we fine-tune the pre-trained backbone using the joint loss $\mathcal{L}_{\lambda = 0.4}$ (Equation~\eqref{eq:joint-loss}) and bias the generative component at inference time using the predictive component of the joint model as the reward. The predictor estimates MIC values for generated peptide sequences, which we normalize into $\hat{\text{MIC}}$ via z-scoring. The reward is defined as $R(\mathbf{x}) = -\hat{\text{MIC}}(\mathbf{x})$, favoring sequences with lower predicted MIC values. We sample until 1{,}000 peptide sequences are collected, without further retraining. Generated peptides are evaluated using the metrics defined above. All experimental details are provided in Appendix~\ref{appendix:amp}.

We compare \textsc{JSI} against established baselines: PepCVAE~\cite{das2018pepcvae}, AMPGAN~\cite{vanoort2021ampganv2}, HydrAMP~\cite{Szymczak2023HydrAMP}, AMP-Diffusion~\cite{chen2023ampdiffusion}, and Hyformer~\cite{izdebski2026synergistic}.

\textsc{JSI} outperforms all baselines on AMP-related oracles: HydrAMP$_{\text{MIC}}$, AMPlify, and AmPEPpy, while maintaining competitive diversity and fitness (Table~\ref{tab:amp_results}). Specifically, while Hyformer achieves the highest fitness score, \textsc{JSI} improves upon Hyformer across all three AMP oracles, indicating stronger alignment with antimicrobial activity objectives. Baselines such as PepCVAE, AMPGAN, and AMP-Diffusion achieve considerably lower performance across all AMP-related metrics. We report the uniqueness and novelty of the generated peptides in Appendix~\ref{appendix:amp-novelty-diversity}, and evaluate \textsc{JSI} under reduced evaluation budgets and offline dataset sizes in Appendix~\ref{appendix:limited-evaluation}.

Under the challenging setting of restricting the proportion of high-scoring peptides, \textsc{JSI} maintains strong performance across all three AMP oracles, outperforming Hyformer by a large margin, notably even in the 0\% setting, where no hits are available during fine-tuning (Table~\ref{tab:amp_results_harder}). While Hyformer maintains higher diversity, \textsc{JSI} consistently achieves superior fitness and AMP oracle scores. The difference is particularly pronounced on HydrAMP$_{\text{MIC}}$, the most challenging oracle, where \textsc{JSI} achieves over threefold improvement over Hyformer in both the 1\% and 0\% settings.

\begin{table*}[ht!]
    \caption{Molecular optimization performance on offline AMP optimization. Best method is \textbf{bold}.}
    \vspace{-2ex}
    \label{tab:amp_results}
    \begin{center}
    \begin{scriptsize}
    \begin{sc}
    \begin{adjustbox}{width=\linewidth}
    \begin{tabular}{@{}lcccccc@{}}
    \toprule
    Model & Perplexity & Diversity ($\uparrow$) & Fitness ($\uparrow$) & HydrAMP$_{\text{MIC}}$ ($\uparrow$) & AMPlify ($\uparrow$) & AmPEPpy ($\uparrow$) \\
    \midrule
    PepCVAE       &  20.39 (20.17, 20.64) & \textbf{0.87 (0.87, 0.87)} & 0.07 (0.07, 0.07)  & 0.12 (0.10, 0.14) & 0.48 (0.46, 0.50) & 0.52 (0.51, 0.53) \\
    AMPGAN        & 18.97 (18.76, 19.19)  & 0.80 (0.80, 0.81) & 0.12 (0.12, 0.13)  & 0.22 (0.20, 0.25) &  0.66 (0.63, 0.68) & 0.55 (0.54, 0.57) \\
    HydrAMP       & 19.98 (19.71, 20.27)  & 0.86 (0.86, 0.86) & 0.09 (0.08, 0.09)  & 0.44 (0.41, 0.47) & 0.60 (0.57, 0.62) & 0.52 (0.52, 0.53) \\
    AMP-Diffusion & 18.33 (18.03, 18.64) & 0.82 (0.82, 0.82) & 0.11 (0.10, 0.11) & 0.18 (0.16, 0.21) & 0.14 (0.12, 0.15)  & 0.33 (0.31, 0.34)\\
    Hyformer      & 17.72 (17.47, 17.97)  &  0.74 (0.73, 0.74)  & \textbf{0.21 (0.20, 0.22)}  & 0.78 (0.76, 0.81) & 0.96 (0.95, 0.97) & 0.71 (0.70, 0.72) \\
    \textsc{JSI (ours)} & 15.93 (15.68, 16.17) & 0.78 (0.78, 0.79)& 0.12 (0.12, 0.12) & \textbf{0.84 (0.82, 0.87)} & \textbf{0.98 (0.97, 0.98)} & \textbf{0.78 (0.77, 0.78)}\\
    \bottomrule
    \end{tabular}
    \end{adjustbox}
    \end{sc}
    \end{scriptsize}
    \end{center}
    \vspace{-2ex}
\end{table*}

\begin{table*}[ht!]
    \caption{Molecular optimization performance on offline AMP optimization with restricted proportion of hits. Best method is \textbf{bold}.}
    \vspace{-2ex}
    \label{tab:amp_results_harder}
    \begin{center}
    \begin{scriptsize}
    \begin{sc}
    \begin{adjustbox}{width=\linewidth}
    \begin{tabular}{@{}lcccccc@{}}
    \toprule
    Model & Perplexity & Diversity ($\uparrow$) & Fitness ($\uparrow$) & HydrAMP$_{\text{MIC}}$ ($\uparrow$) & AMPlify ($\uparrow$) & AmPEPpy ($\uparrow$) \\
    \midrule
    \textbf{1\% hits} \\
    \quad Hyformer & 16.84 (16.59, 17.09) & \textbf{0.83 (0.82, 0.83)} & 0.11 (0.11, 0.12) & 0.20 (0.18, 0.23) & 0.76 (0.74, 0.78) & 0.65 (0.64, 0.66)\\
    \quad \textsc{JSI (ours)} & 18.36 (17.94, 18.77)& 0.69 (0.68, 0.70) & \textbf{0.16 (0.16, 0.17)} & \textbf{0.68 (0.65, 0.71)} & \textbf{0.98 (0.97, 0.98)} & \textbf{0.72 (0.72, 0.73)} \\
    \midrule
    \textbf{0\% hits} \\
    \quad Hyformer  & 16.77 (16.51, 17.01) & \textbf{0.83 (0.83, 0.83)} & 0.11 (0.10, 0.11) &  0.17 (0.15, 0.19) & 0.71 (0.69, 0.73) & 0.62 (0.61, 0.63)\\
    \quad \textsc{JSI (ours)} & 18.34 (18.05, 18.66) & 0.72 (0.72, 0.73) & \textbf{0.19 (0.18, 0.20)} & \textbf{0.66 (0.63, 0.69)} & \textbf{0.98 (0.97, 0.99)} & \textbf{0.71 (0.70, 0.72)}\\
    \bottomrule
    \end{tabular}
    \end{adjustbox}
    \end{sc}
    \end{scriptsize}
    \end{center}
    \vspace{-2ex}
\end{table*}

\subsection{Ablations}\label{section:ablations}

To quantify the individual contributions of joint learning and inference-time self-improvement, we ablate \textsc{JSI} on sample-efficient offline docking score optimization in two variants: (i) \textsc{w/o joint modeling}, which decouples generation and prediction by using the offline dataset to separately fine-tune the generative part of the joint model and a standalone predictor, and (ii) \textsc{w/o inference-time self-improvement (si)}, which reduces our sampling procedure to Best-of-K sampling~\cite{gao2023scaling, beirami2025theoretical}, without replacement. For the predictor, we use Chemprop~\cite{heid2023chemprop}, a state-of-the-art message-passing neural network for molecular property prediction. For IntDiv1 in Table~\ref{table:ablations}, we report the mean across 10 random seeds, aggregated over all protein targets. All experiments are conducted in the offline docking score optimization setting, with all details in Appendix~\ref{appendix:ablations}. 

\textsc{JSI} significantly outperforms both ablated variants on all protein targets (two-sided Welch's t-test, $p<0.05$), except on 5HT1B. We hypothesize that 5HT1B remains the easiest protein target, as compounds with high binding affinity are relatively common in the ZINC dataset (see Table~\ref{experiments:table:hit-ratio-online}), which explains the relatively high performance of the ablated versions. As expected, the variant without inference-time \textsc{SI} attains a slightly higher diversity.

\begin{table*}[ht!]
    \caption{Ablations for molecular optimization performance on docking score optimization in the offline setting. We mark the best method \textbf{bold}. \textsc{SI} - self improvement.}
    \vspace{-2ex}
    \label{table:ablations}
    \begin{center}
    \begin{scriptsize}
    \begin{sc}
    \begin{adjustbox}{width=\linewidth}
    \begin{tabular}{@{}lcccccc@{}}
    \toprule
     & \multicolumn{5}{c}{Target Protein, Hit Ratio ($\uparrow$)} & \\ 
    \cmidrule{2-6}
    Method       & parp1 & fa7 & 5ht1b & braf & jak2 & IntDiv1 ($\uparrow$) \\
    \midrule
    \textsc{JSI} & \textbf{45.628 (11.653)} & \textbf{5.831 (2.040)} & 76.940 (4.442) & \textbf{21.637 (4.607)} & \textbf{42.001 (4.344)} & 0.763 (0.015) \\
    \quad \textsc{w/o joint modeling} & 25.309 (8.673) & 0.479 (0.771) & \textbf{77.579 (4.343)} & 7.298 (3.295) & 13.333 (6.047) & 0.713 (0.043) \\
    \quad \textsc{w/o inference-time si} & 29.468 (4.059) & 2.629 (0.619) & 73.347 (2.131) & 10.978 (1.936) & 29.432 (2.346) & \textbf{0.826 (0.009)} \\
    \bottomrule
  \end{tabular}
  \end{adjustbox}
  \end{sc}
  \end{scriptsize}
  \end{center}
  \vspace{-4ex}
\end{table*}

\section{Conclusions}\label{section:conclusions}
In this paper, we introduced \textsc{JSI}, a framework for sample-efficient molecular optimization in both online and offline settings. By combining a joint generative-predictive model with inference-time self-improvement, \textsc{JSI} successfully improves sample efficiency, providing state-of-the-art performance in docking score optimization across both offline and online settings, and in AMP optimization. Notably, \textsc{JSI} generates optimized peptides even when no high-scoring examples are present in the offline data.

\paragraph{Limitations \& Future Work} In challenging tasks, inaccuracies in the surrogate component may limit optimization performance. While the surrogate correlates strongly with the oracle on held-out molecules, this correlation degrades on the generated molecules. As sampling is performed at inference-time, rather than through iterative fine-tuning, inaccuracies do not accumulate, yet future work should explore methods to further improve the robustness of surrogates in the offline setting. Moreover, inference-time self-improvement trades off diversity for optimization performance, and explicitly controlling this trade-off remains a challenge.
Despite these limitations, \textsc{JSI} outperforms state-of-the-art approaches in sample-efficient molecular optimization across both offline and online settings, with ablations confirming the individual contributions of joint modeling and inference-time self-improvement.

\section{Acknowledgments and Disclosures}

\begin{wrapfigure}{r}{0.3\linewidth}
    \centering
    \vspace{-6ex}
    \includegraphics[width=\linewidth]{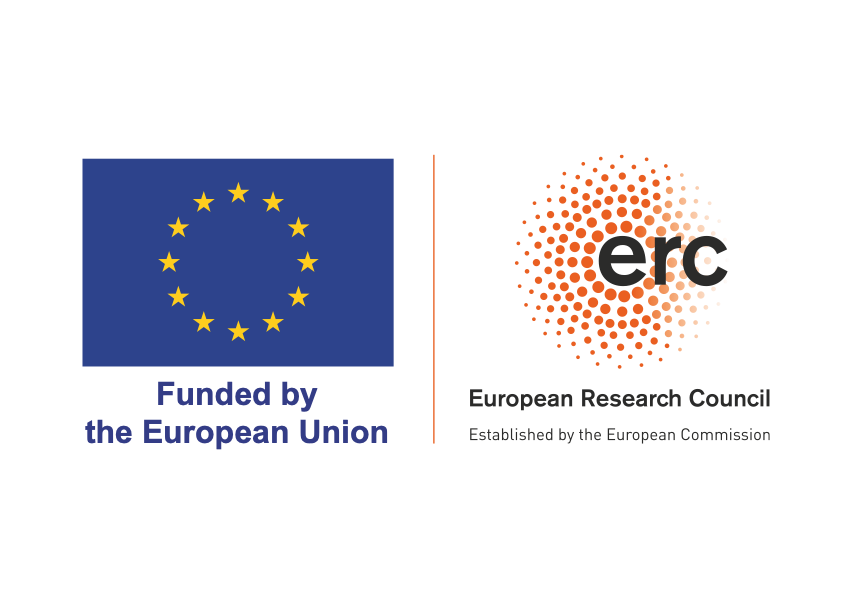}
    \vspace{-8ex}
\end{wrapfigure}
We thank Irene Bonafonte-Pardàs for her valuable comments that improved the final version of the manuscript. This work was supported by Helmholtz Munich's Health AI Fellowship to Adam Izdebski, ERC grant DOG-AMP (No 101125506), funded by the European Union, and the Deutsche Forschungsgemeinschaft (DFG, German Research Foundation) through Project No. 466387255 within the Priority Programme “SPP 2331: Machine Learning in Chemical Engineering” (JP, DGG). Views and opinions expressed are however those of the author(s) only and do not necessarily reflect those of the European Union or the European Research Council Executive Agency. Neither the European Union nor the granting authority can be held responsible for them. We acknowledge the use of the HPC cluster at Helmholtz Munich and of generative AI tools to improve the final version of the manuscript. All AI-generated content was reviewed, edited and approved by the authors.  

\newpage
\bibliographystyle{abbrvnat}
\bibliography{main}

\appendix

\section{Impact Statement}
This work advances deep generative modeling for molecular optimization, with direct applications to drug discovery. The primary societal benefit is accelerating the design of novel therapeutics, particularly for indications where experimental screening is expensive or slow. A potential risk of generative molecular optimization is dual-use: the same methods could in principle be applied to design harmful substances. We encourage responsible use of these tools within established ethical and regulatory frameworks governing chemical and biological research.

\section{Notation}\label{appendix:mathematical-notation} We follow the mathematical notation of~\citet{goodfellow2016book}. Table~\ref{tab:notation} summarizes the symbols used throughout the paper.

\begin{table}[!h]
    \caption{Mathematical notation used throughout the paper.}
    \label{tab:notation}
    \begin{center}
    \begin{tabular}{@{}cl@{}}
    \toprule
        \textbf{Symbol} & \textbf{Meaning} \\ 
    \midrule
        $[N]$  & Set of integers $1, \ldots, N$ \\
        $\mathbf{a}$    & Vector (column-vector) \\
        $\mathbf{a}^{(i)}$    & Vector indexed for some purpose \\
        $\mathbf{a}_{i:j}$ & Vector with values from the $i$-th to $j$-th (inclusive) entry of vector $\mathbf{a}$ \\
        $a_{i}$ & The $i$-th entry of vector $\mathbf{a}$ \\
        $a$ & Scalar \\ 
        $\X$   & Input space, i.e.\ the space of all possible inputs \\
        $\mathcal{Y}$   & Target space, i.e.\ the space of all possible property values \\
        $\mathcal{V}$ & Vocabulary, i.e.\ the set of tokens \\
        $\mathcal{D}$ & Dataset \\
        $p_\theta(\x, y)$ & Joint model parametrized with parameters $\theta \in \Theta$ \\
        $p_\theta(y \mid \x)$ & Predictive model parametrized with parameters $\theta \in \Theta$ \\
        $p_\theta(\x)$ & Generative model parametrized with parameters $\theta \in \Theta$ \\
    \bottomrule
    \end{tabular}
    \end{center}
\end{table}

\section{Theory and Derivations}\label{appendix:theory}

\subsection{Derivation of the Perturbed Logits}\label{appendix:perturbed-logits}
We derive the perturbed logits in Eq.~\eqref{eq:perturbed-model-practical}, showing that they combine sampling without replacement with a reward-dependent exponential tilt, and the subtracted probability term is the probability mass remaining under a prefix after previously selected complete sequences have been removed.

Let $u := \x_{1:t-1}$ denote the current parent prefix and let $a := x_t \in \mathcal V$ be a candidate next token from the vocabulary $\mathcal V$, and write $ua$ for their concatenation. Let $\mathcal C(ua)=\{\x : \x_{1:t}=ua\}$ be the set of all complete sequences extending $ua$. For an autoregressive model
\begin{equation}
    p_\theta(a\mid u) = \sum_{\x\in\mathcal C(ua)}p_\theta(\x\mid u).
\end{equation}
Let $\mathcal S=\{\x^{(k)}\}_{k=1}^K$ be the set of complete sequences sampled so far, and define $S(ua)=\mathcal S\cap\mathcal C(ua)$. The quantity inside the logarithm in Eq.~\eqref{eq:perturbed-model-practical} is then
\begin{equation}
    r_u(a) := p_\theta(a\mid u) - \sum_{\x\in S(ua)}p_\theta(\x\mid u) = \sum_{\x\in\mathcal C(ua)\setminus\mathcal S}p_\theta(\x\mid u),
\end{equation}
which immediately gives $r_u(a)\geq 0$. Moreover, $r_u(a)>0$ whenever at least one unsampled completion of $ua$ has positive model probability. The term is therefore strictly positive for every \emph{active} (i.e., visitable) branch of the trie. It is exactly zero only when the branch is exhausted (its full subtree has been sampled), in which case the logit is $\log(0)=-\infty$ and the branch cannot be selected again. This construction yields sampling without replacement. At a non-exhausted parent prefix $u$, the total residual mass is
\begin{equation}
    Z_u = \sum_{a\in\mathcal V} r_u(a) = 1-\sum_{\x\in\mathcal S\cap\mathcal C(u)}p_\theta(\x\mid u) > 0,
\end{equation}
and the normalized residual mass is
\begin{equation}
    \frac{r_u(a)}{Z_u} = \Pr\nolimits_\theta\!\left(X_t = a \mid X_{1:t-1} = u,\ X \notin \mathcal S\right).
\end{equation}
The ``subtract-and-renormalize'' operation is hence the original model conditioned on not returning a previously sampled sequence. This is the standard incremental sampling-without-replacement interpretation~\citep{kool2019stochastic, pirnay2024selfimprovement}: after a leaf is sampled, its mass is recursively removed from that leaf and all its ancestors, and the resulting trie represents the model conditioned on that leaf being unavailable.

\subsection{Link to logit-tilted sampling}
Let $A_u(a) = \sum_{\x\in S(ua)}\textrm{Adv}_\theta(\x)$. After transforming the logits to probabilities, Eq.~\eqref{eq:perturbed-model-practical} is equivalent to
\begin{equation}
    p_{\theta'}(a\mid u,\mathcal S) = \frac{r_u(a)\exp\bigl(\sigma A_u(a)\bigr)}{\sum_{b\in\mathcal V} r_u(b)\exp\bigl(\sigma A_u(b)\bigr)}.
\end{equation}
This makes the two operations explicit:
\begin{enumerate}
    \item $r_u(a)$ removes the mass of already sampled sequences and therefore implements sampling without replacement.
    \item $\exp(\sigma A_u(a))$ tilts the remaining mass toward prefixes associated with higher reward.
\end{enumerate}
Since $\exp(\sigma A_u(a))>0$, we have $p_{\theta'}(a\mid u,\mathcal S)=0$ if and only if $r_u(a)=0$. The tilt therefore cannot assign mass to a continuation whose completions have all been sampled.

\subsection{Connection to the REINVENT Algorithm}\label{appendix:REINVENT-gradient}
REINVENT~\cite{loeffler2024reinvent} fine-tunes the model $\generative$ by minimizing the squared difference between $\log\generative$ and the log-likelihood of an augmented distribution $p_{\textrm{aug}}(\x) \propto \reference\exp\bigl(\sigma f(\x)\bigr)$, where $\sigma = \beta^{-1}$ is a scaling parameter. For $\x \sim \generative$, the gradient of the REINVENT loss coincides with the summand of the Monte Carlo estimator in Eq.~\eqref{eq:mc-gradient} up to a constant $c \in \R$, as
\begin{equation}
    \begin{split}
        \mathcal{L}_{\textrm{REINVENT}}(\theta) &= \Bigl(\log p_{\textrm{aug}}(\x) - \log p_\theta(\x)\Bigr)^2\\
        &= \Bigl(\log \reference + \sigma f(\x) - \log p_\theta(\x)\Bigr)^2 \\
        &= \Bigl(\sigma f(\x) - \log \frac{p_\theta(\x)}{\reference}\Bigr)^2 \\
        &= \sigma^2 \Bigl(f(\x) - \frac{1}{\sigma}\log \frac{p_\theta(\x)}{\reference}\Bigr)^2
    \end{split}
\end{equation}
and
\begin{align}
    \nabla_\theta \mathcal{L}_\textrm{REINVENT}(\theta) &= \nabla_\theta\sigma^2 \Bigl(f(\x) - \frac{1}{\sigma}\log \frac{p_\theta(\x)}{\reference}\Bigr)^2 \nonumber \\
    &= c\Bigl(f(\x) - \beta\log \frac{p_\theta(\x)}{\reference}\Bigr)\nabla_\theta\log p_\theta(\x),
\end{align}
with $c = -2\sigma$. As $\x$ is sampled from $\generative$, REINVENT-style algorithms inherit the sample inefficiency of Eq.~\eqref{eq:mc-gradient}, i.e., when high-scoring molecules are rare under $\generative$, most terms of the gradient carry little signal about $f$.

\section{Algorithm and Implementation of \textsc{JSI}}\label{appendix:sampling-with-jsi}

We provide the full sampling procedure for \textsc{JSI} in the offline setting in Algorithm~\ref{alg:joint-self-improvement}. At each round, \textsc{JSI} draws $K$ candidate sequences using SBS (Line~\ref{step:sbs}), scores them via the predictive head to obtain the expected prediction score $\mu$ (Line~\ref{step:mu}) and the global advantages (Line~\ref{step:advantage}), and biases the next-token probabilities towards higher-scoring continuations by adding the scaled advantage to the logits (Line~\ref{step:logits}, Eq.~\eqref{eq:perturbed-model-practical}). The perturbed logits are then normalized with a softmax to yield the model used in the next round (Line~\ref{step:softmax}). This is repeated for $N_{\textrm{rounds}}$ iterations, progressively steering generation towards high-scoring molecules, and the highest-scoring sampled sequence is returned (Line~\ref{step:selection}). In the online setting, the predicted score $p_\theta(y \mid \x)$ is replaced by oracle evaluations $f(\x)$. In practice, the offline predictions and the online evaluations are further aggregated into the rewards $\hat{R}(\x)$ defined in Eq.~\eqref{eq:reward-offline} and $R(\x)$ in Eq.~\eqref{eq:reward-online}, respectively.

\begin{algorithm}[t]
\caption{Sampling with \textsc{JSI} in the offline setting}
\label{alg:joint-self-improvement}
\begin{algorithmic}[1]
\Require Joint model $p_\theta(\mathbf{x}, y)$, number of rounds $N_{\textrm{rounds}}$, beam width $K$, step size $\sigma$
\Ensure Sample $\mathbf{x}$ with highest $p_\theta(y \mid \mathbf{x})$
\State $\mathcal{D}_\textrm{sampled} \leftarrow \emptyset$
\For{$n = 1,\ldots,N_\textrm{rounds}$}
    \State $\{\mathbf{x}^{(k)}\}_{k=1}^K \leftarrow \text{SBS}(p_{\theta}(\x), K)$ \label{step:sbs}
    \State $\mathcal{D}_\textrm{sampled} \leftarrow \mathcal{D}_\textrm{sampled} \cup \{\mathbf{x}^{(k)}\}_{k=1}^K$
    \State $\mu \leftarrow \mu\bigl(\mathbf{x}^{(1)},\ldots,\mathbf{x}^{(K)}\bigr)$ \label{step:mu}
    \State $p_{\theta'}(\x) \leftarrow p_\theta(\x)$
    \For{$k = 1,\ldots,K$}
        \For{$t = 1,\ldots,T$}
            \State $A_t^{(k)} \leftarrow
                \displaystyle\sum_{\mathbf{x} \in S(\mathbf{x}^{(k)}_{1:t})}
                \Bigl(p_\theta(y \mid \mathbf{x}) - \mu\Bigr)$ \label{step:advantage}
            \State $\pi_t^{(k)} \leftarrow
                \displaystyle\sum_{\mathbf{x} \in S(\mathbf{x}^{(k)}_{1:t})}
                p_\theta\bigl(\mathbf{x} \mid \mathbf{x}_{1:t-1}^{(k)}\bigr)$
            \State $\logit\, p_{\theta'}\bigl(x^{(k)}_t \mid \mathbf{x}^{(k)}_{1:t-1}\bigr) \leftarrow \log\Bigl(p_\theta\bigl(x^{(k)}_t \mid \mathbf{x}^{(k)}_{1:t-1}\bigr) - \pi_t^{(k)}\Bigr) +\sigma \cdot A_t^{(k)}$ \label{step:logits}
            \hfill\textup{(Eq.~\eqref{eq:perturbed-model-practical})}
        \EndFor
    \EndFor
    \State $p_{\theta}(\x) \leftarrow \mathrm{softmax}\bigl(\logit\, p_{\theta'}(\x)\bigr)$ \label{step:softmax}
\EndFor
\State \Return $\argmax_{\mathbf{x} \in \mathcal{D}_\textrm{sampled}} p_\theta(y \mid \mathbf{x})$ \label{step:selection}
\end{algorithmic}
\end{algorithm}

\subsection{Numerical Stability in Eq.~\eqref{eq:perturbed-model-practical}}\label{appendix:numerical-stability}
We carry out all operations in log space and store subtree probability masses as log masses. Specifically, if $x$ and $y$ are log probabilities with $x \geq y$, we never form the potentially unstable difference $\exp(x)-\exp(y)$ directly. Instead, with $x$ the log of the currently remaining child mass and $y$ the log mass to be removed, we use the numerically stable form (computed with \texttt{np.log1p}, in the spirit of the log-sum-exp trick)
\begin{equation}
    \operatorname{logsubexp}(x,y) := \log\bigl(\exp(x) - \exp(y)\bigr) = x+\log\bigl(1-\exp(y-x)\bigr).
\end{equation}
The trie update uses three further safeguards. First, exhausted leaves or subtrees are explicitly assigned an unsampled log mass of $-\infty$, which prevents them from being sampled again. Second, for log masses $x, y, z$,
\begin{equation}
    \log\bigl(\exp(x) - \exp(y) - \exp(z)\bigr) = \log\bigl(\exp(\operatorname{logsubexp}(x,y)) - \exp(z)\bigr),
\end{equation}
so instead of evaluating the sum of many small linear probabilities in Eq.~\eqref{eq:perturbed-model-practical}, we incrementally subtract the log mass of each sampled leaf from all its ancestors. Finally, the advantages are added to the residual log masses, after which normalization uses the standard max-shifted exponentiation $\exp(\log a_i- \max_j \log a_j)$, which avoids overflow. An exhausted node is detected by checking that all of its child log masses are $-\infty$, and such a node is not expanded further.

\section{Additional Experiments}\label{appendix:additional-experiments}

\subsection{Reward Normalization and Aggregation}\label{appendix:reward-analysis}
To assess the effect of reward normalization and aggregation, we compare three combinations of normalization and aggregation on the BRAF protein target, using 10 random seeds in the online docking score optimization setting (Section~\ref{section:online-molecular-optimization}). Let $v(\x)$ be the raw value of property $v \in \{\mathrm{DS}, \mathrm{SA}, \mathrm{QED}\}$, and consider Z-score normalization, as in Eq.~\eqref{eq:reward-online},
\begin{equation}\label{eq:reward-normalization}
    v^{\textrm{z}}(\x) = \frac{v(\x) - \mu_v}{\sigma_v},
\end{equation}
where $\mu_v$ and $\sigma_v$ are computed from the oracle evaluations available at the current optimization round, as well as the min-max normalization $v^{\textrm{mm}}$ of SATURN~\cite{guo2024saturnsampleefficientgenerativemolecular}, which maps each property into $[0, 1]$ and which we take from their implementation. Next, let $s_v(\x)$ be the normalized value of property $v$, so that $s_v(\x)$ is either $v^{\textrm{z}}(\x)$ or $v^{\textrm{mm}}(\x)$, oriented so that higher is better, since lower values of DS and SA are preferred, and consider a sum and a product
\begin{equation}\label{eq:reward-aggregation}
    R^{\textrm{sum}}(\x) = \frac{1}{3}\sum_v s_v(\x), \qquad R^{\textrm{prod}}(\x) = \prod_v s_v(\x),
\end{equation}
so that Z-score normalization with sum aggregation recovers the reward in Eq.~\eqref{eq:reward-online}, whereas min-max normalization with product aggregation is the reward of SATURN. Product aggregation requires non-negative scores and is therefore paired with min-max normalization. Z-score normalization with sum aggregation, which is JSI's default, achieves the highest Hit Ratio (Table~\ref{table:reward_analysis}), outperforming both min-max normalization with sum aggregation and min-max normalization with product aggregation.

\begin{table*}[h]
    \caption{Hit Ratio for different reward normalization and aggregation methods in the online docking score optimization setting. We mark the best configuration \textbf{bold}. Values are mean (standard deviation) across 10 random seeds.}
    \begin{center}
    \begin{scriptsize}
    \begin{sc}
    \begin{tabular}{@{}llc@{}}
    \toprule
    & & \multicolumn{1}{c}{Target Protein, Hit Ratio ($\uparrow$)} \\
    \cmidrule{3-3}
    Normalization & Aggregation & braf \\
    \midrule
    Z-scoring & Sum     & \textbf{58.547 (11.456)} \\
    Min-max   & Sum     & 23.223 (6.426) \\
    Min-max   & Product & 38.897 (7.024) \\
    \bottomrule
    \end{tabular}
    \end{sc}
    \end{scriptsize}
    \end{center}
    \label{table:reward_analysis}
\end{table*}

\subsubsection{Weighted Reward}\label{appendix:weighted-reward}
To assess the effect of the relative weighting of individual properties, we generalize the reward in Eq.~\eqref{eq:reward-online} to the weighted reward
\begin{equation}\label{eq:reward-online-weighted}
    R_{\mathbf{w}}(\x) = w_{\mathrm{DS}}\bigl(-\mathrm{DS}^{\textrm{z}}(\x)\bigr) + w_{\mathrm{SA}}\bigl(-\mathrm{SA}^{\textrm{z}}(\x)\bigr) + w_{\mathrm{QED}}\,\mathrm{QED}^{\textrm{z}}(\x),
\end{equation}
i.e. Z-score normalization with sum aggregation (Eq.~\eqref{eq:reward-aggregation}), where $\mathbf{w} = (w_{\mathrm{DS}}, w_{\mathrm{SA}}, w_{\mathrm{QED}})$, with $w_i \geq 0$ and $\sum_i w_i = 1$, so that the uniform weighting $\mathbf{w} = (1/3, 1/3, 1/3)$ recovers Eq.~\eqref{eq:reward-online}. We vary $\mathbf{w}$ on all five protein targets in the online docking score optimization setting (Section~\ref{section:online-molecular-optimization}), using 10 random seeds (Table~\ref{table:weighted-reward}). Downweighting the docking score to 0.25 lowers the Hit Ratio on all five targets relative to uniform weighting. Upweighting the docking score to 0.5 improves the Hit Ratio on four of the five targets, namely PARP1, FA7, BRAF and JAK2, with 5HT1B the only exception. Increasing the docking-score weight further to 0.75 does not help: the Hit Ratio falls below that of the 0.5 setting on every target and is comparable to or below that of uniform weighting on all targets except FA7. These results indicate that moderately increasing the relative weight of the docking score can improve performance on most targets.

\begin{table*}[h]
    \caption{Hit Ratio for different weightings $\mathbf{w}$ of the reward $R_{\mathbf{w}}$ (Eq.~\eqref{eq:reward-online-weighted}) in the online docking score optimization setting. We mark the best weighting per target \textbf{bold}. Values are mean (standard deviation) across 10 random seeds.}
    \begin{center}
    \begin{scriptsize}
    \begin{sc}
    \begin{tabular}{@{}lccccc@{}}
    \toprule
    & \multicolumn{5}{c}{Target Protein, Hit Ratio ($\uparrow$)} \\
    \cmidrule{2-6}
    $w_{\mathrm{DS}}$ / $w_{\mathrm{SA}}$ / $w_{\mathrm{QED}}$ & parp1 & fa7 & 5ht1b & braf & jak2 \\
    \midrule
    $0.25$ / $0.375$ / $0.375$ & 65.163 (7.858) & 34.350 (11.492) & 87.827 (3.689) & 49.883 (10.589) & 68.553 (12.792) \\
    $0.33$ / $0.33$ / $0.33$ & 67.280 (7.961)  & 36.587 (10.270) & \textbf{91.357 (3.204)} & 58.547 (11.456) & 71.863 (6.127) \\
    $0.5$ / $0.25$ / $0.25$ & \textbf{75.470 (6.973)} & \textbf{43.530 (11.954)} & 84.413 (6.580) & \textbf{64.997 (7.474)} & \textbf{80.230 (4.915)} \\
    $0.75$ / $0.125$ / $0.125$ & 68.913 (9.576) & 41.317 (10.804) & 78.410 (4.293) & 59.570 (9.114) & 67.867 (5.982) \\
    \bottomrule
    \end{tabular}
    \end{sc}
    \end{scriptsize}
    \end{center}
    \label{table:weighted-reward}
\end{table*}

\subsection{Surrogate Accuracy}

To evaluate the accuracy of the predictive part of the joint model, we compare its predictions against ground-truth oracle docking scores across all protein targets, using the Pearson correlation, on both held-out sequences (Test) and sequences produced in the offline docking score optimization setting (Generated), reported in Section~\ref{section:offline-molecular-optimization}. The held-out dataset consists of 500 molecules randomly selected from ZINC250K and kept fixed across random seeds and protein targets, for which we compute the corresponding docking score, SA and QED for each target. We report the mean and standard deviation calculated across 10 random seeds (Table~\ref{tab:surrogate-accuracy}).

\begin{table*}[h]
    \caption{Surrogate predictive performance on held-out (Test) and generated (Generated) sequences in the offline docking score optimization setting, measured by the Pearson correlation with the ground-truth oracle docking score. Values are mean (standard deviation) across 10 random seeds.}
    \begin{center}
    \begin{scriptsize}
    \begin{sc}
    \begin{tabular}{@{}lccccc@{}}
    \toprule
    & \multicolumn{5}{c}{Target Protein, Pearson Correlation ($\uparrow$)} \\
    \cmidrule{2-6}
    Sequences & parp1 & fa7 & 5ht1b & braf & jak2 \\
    \midrule
    Test      & 0.879 (0.005) & 0.762 (0.012) & 0.867 (0.007) & 0.802 (0.010) & 0.832 (0.006) \\
    Generated & 0.224 (0.190) & 0.284 (0.150) & 0.305 (0.194) & 0.304 (0.141) & 0.384 (0.122) \\
    \bottomrule
    \end{tabular}
    \end{sc}
    \end{scriptsize}
    \end{center}
    \label{tab:surrogate-accuracy}
\end{table*}

On the held-out sequences, the surrogate maintains a strong positive correlation with the oracle values on all five targets, confirming its strong predictive performance. On the generated sequences, the correlation remains positive, although with a markedly higher variance across seeds, indicating that the surrogate retains sufficient signal to prioritize high-scoring candidates during optimization. A lower correlation is expected here, as \textsc{JSI} generates high-scoring molecules that may be increasingly distinct from the offline dataset.

\subsection{Hyperparameter Sensitivity}
To assess the sensitivity of \textsc{JSI} to its hyperparameters, we vary the step size $\sigma$, the beam width $K$ and the joint-loss weight $\lambda$ on the BRAF protein target in the offline docking score optimization setting (Section~\ref{section:offline-molecular-optimization}), using 3 random seeds. Across all values of $\sigma$ and $K$, the Hit Ratio stays within one standard deviation of the default configuration ($\sigma = 1$, $K = 128$, $\lambda = 0.25$), indicating robustness to the exact choice of these hyperparameters (Table~\ref{tab:sens-advantage}). Moreover, the Hit Ratio increases monotonically with the beam width (Table~\ref{tab:sens-beam}). Finally, values of $\lambda \geq 0.4$ degrade performance (Table~\ref{tab:sens-lm}).

\begin{table}[h]
    \caption{Sensitivity analysis for the step size $\sigma$ in the offline docking score optimization setting. Values are mean (standard deviation) across 3 random seeds.}
    \begin{center}
    \begin{scriptsize}
    \begin{sc}
    \begin{tabular}{@{}lc@{}}
    \toprule
    & \multicolumn{1}{c}{Target Protein, Hit Ratio ($\uparrow$)} \\
    \cmidrule{2-2}
    Step Size $\sigma$ & braf \\
    \midrule
    0.25 & 33.829 (14.137) \\
    0.5  & 29.907 (10.995) \\
    0.75 & 27.402 (7.025) \\
    1    & 27.418 (9.888) \\
    \bottomrule
    \end{tabular}
    \end{sc}
    \end{scriptsize}
    \end{center}
    \label{tab:sens-advantage}
\end{table}

\begin{table}[h]
    \caption{Sensitivity analysis for the beam width $K$ in the offline docking score optimization setting. Values are mean (standard deviation) across 3 random seeds.}
    \begin{center}
    \begin{scriptsize}
    \begin{sc}
    \begin{tabular}{@{}lc@{}}
    \toprule
    & \multicolumn{1}{c}{Target Protein, Hit Ratio ($\uparrow$)} \\
    \cmidrule{2-2}
    Beam Width $K$ & braf \\
    \midrule
    16  & 18.414 (3.963) \\
    32  & 21.275 (9.278) \\
    64  & 21.857 (7.136) \\
    128 & 27.418 (9.888) \\
    \bottomrule
    \end{tabular}
    \end{sc}
    \end{scriptsize}
    \end{center}
    \label{tab:sens-beam}
\end{table}

\begin{table}[h]
    \caption{Sensitivity analysis for the joint-loss weight $\lambda$ in the offline docking score optimization setting. Values are mean (standard deviation) across 3 random seeds.}
    \begin{center}
    \begin{scriptsize}
    \begin{sc}
    \begin{tabular}{@{}lc@{}}
    \toprule
    & \multicolumn{1}{c}{Target Protein, Hit Ratio ($\uparrow$)} \\
    \cmidrule{2-2}
    Joint-Loss Weight $\lambda$ & braf \\
    \midrule
    0.1  & 24.410 (8.217) \\
    0.25 & 27.418 (9.888) \\
    0.4  & 18.867 (2.780) \\
    0.5  & 16.564 (2.297) \\
    \bottomrule
    \end{tabular}
    \end{sc}
    \end{scriptsize}
    \end{center}
    \label{tab:sens-lm}
\end{table}

\subsection{Novelty and Diversity for AMP Sequences}\label{appendix:amp-novelty-diversity}
To assess the novelty and diversity of the generated AMP sequences, we extend the evaluation of the AMP sequences generated in
Table~\ref{tab:amp_results} with three metrics computed using the seqme benchmarking
library~\cite{moller2026seqme}. We measure \textbf{Uniqueness}, defined as the fraction
of generated sequences that are unique; \textbf{Novelty}, defined as the fraction of
generated sequences that do not appear in the training data; and the \textbf{3-gram
Jaccard similarity}, defined as the average Jaccard similarity between the 3-grams of
each generated sequence and a reference corpus constructed from the 3-grams of the
training data. In addition, we report \textbf{Diversity}, as defined in
Appendix~\ref{appendix:amp-metrics}. We evaluate the best-performing methods, namely
AMP-Diffusion, Hyformer and \textsc{JSI}, on the same 1{,}000 generated sequences per
method as in Section~\ref{section:amp}.

\textsc{JSI} generates exclusively unique sequences and attains a diversity between
that of the two baselines, while its 3-gram composition remains almost disjoint from
the training data (Table~\ref{tab:amp-novelty-diversity}). Its novelty is slightly
lower than that of the baselines, indicating that a small fraction of the generated
sequences is recovered from the training data.

\begin{table*}[h]
    \caption{Uniqueness, novelty and diversity of the AMP sequences generated in the offline AMP optimization setting. Values are computed on the 1{,}000 sequences per method evaluated in Table~\ref{tab:amp_results}.}
    \begin{center}
    \begin{scriptsize}
    \begin{sc}
    \begin{tabular}{@{}lcccc@{}}
    \toprule
    Method & Uniqueness ($\uparrow$) & Novelty ($\uparrow$) & 3-gram Jaccard Similarity & Diversity ($\uparrow$) \\
    \midrule
    AMP-Diffusion       & 0.960 & 1.000 & 0.003 & 0.820 \\
    Hyformer            & 1.000 & 0.999 & 0.002 & 0.740 \\
    \textsc{JSI (ours)} & 1.000 & 0.916 & 0.004 & 0.780 \\
    \bottomrule
    \end{tabular}
    \end{sc}
    \end{scriptsize}
    \end{center}
    \label{tab:amp-novelty-diversity}
\end{table*}

\subsection{Reduced Evaluation Budget and Offline Dataset Size}\label{appendix:limited-evaluation}
To examine whether joint modeling remains applicable under the tighter constraints of
real-world applications, we extend the AMP optimization experiment
(Section~\ref{section:amp}) with two settings. First, we reduce the evaluation budget
below the 1{,}000 sequences of the main experiment
(Table~\ref{tab:eval-budget}). Second, we reduce the number of labeled sequences
available for fine-tuning (Table~\ref{tab:dataset-size}). In both settings we report
the metrics defined in Appendix~\ref{appendix:amp-metrics}.

\textsc{JSI} maintains strong performance across all evaluation budgets, and across all
offline dataset sizes except the smallest, where the scores on all three AMP oracles
degrade, most markedly on HydrAMP$_{\text{MIC}}$, the most challenging oracle.

\begin{table*}[h!]
    \caption{Performance of \textsc{JSI} on offline AMP optimization under a reduced evaluation budget. Values are mean (95\% confidence interval).}
    \begin{center}
    \begin{scriptsize}
    \begin{sc}
    \begin{adjustbox}{width=\linewidth}
    \begin{tabular}{@{}lcccccc@{}}
    \toprule
    Evaluation Budget & Perplexity & Diversity ($\uparrow$) & Fitness ($\uparrow$) & HydrAMP$_{\text{MIC}}$ ($\uparrow$) & AMPlify ($\uparrow$) & AmPEPpy ($\uparrow$) \\
    \midrule
    250 & 16.38 (15.97, 16.83) & 0.76 (0.74, 0.77) & 0.13 (0.12, 0.14) & 0.90 (0.85, 0.93) & 0.98 (0.97, 0.99) & 0.78 (0.77, 0.79) \\
    500 & 15.99 (15.64, 16.35) & 0.77 (0.76, 0.77) & 0.13 (0.12, 0.13) & 0.86 (0.83, 0.89) & 0.98 (0.97, 0.99) & 0.78 (0.77, 0.79) \\
    750 & 16.08 (15.81, 16.33) & 0.78 (0.77, 0.78) & 0.12 (0.11, 0.13) & 0.87 (0.84, 0.89) & 0.98 (0.98, 0.99) & 0.78 (0.77, 0.79) \\
    \bottomrule
    \end{tabular}
    \end{adjustbox}
    \end{sc}
    \end{scriptsize}
    \end{center}
    \label{tab:eval-budget}
\end{table*}

\begin{table*}[h!]
    \caption{Performance of \textsc{JSI} on offline AMP optimization under a reduced offline dataset size. Values are mean (95\% confidence interval).}
    \begin{center}
    \begin{scriptsize}
    \begin{sc}
    \begin{adjustbox}{width=\linewidth}
    \begin{tabular}{@{}lcccccc@{}}
    \toprule
    Offline Dataset Size & Perplexity & Diversity ($\uparrow$) & Fitness ($\uparrow$) & HydrAMP$_{\text{MIC}}$ ($\uparrow$) & AMPlify ($\uparrow$) & AmPEPpy ($\uparrow$) \\
    \midrule
    250 & 17.58 (17.26, 17.91) & 0.79 (0.78, 0.79) & 0.15 (0.14, 0.15) & 0.73 (0.70, 0.75) & 0.91 (0.90, 0.92) & 0.70 (0.69, 0.71) \\
    500 & 15.41 (15.14, 15.65) & 0.70 (0.69, 0.71) & 0.20 (0.19, 0.21) & 0.92 (0.90, 0.93) & 0.98 (0.98, 0.99) & 0.72 (0.72, 0.73) \\
    750 & 13.79 (13.53, 14.08) & 0.76 (0.76, 0.76) & 0.19 (0.18, 0.20) & 0.91 (0.89, 0.92) & 0.98 (0.97, 0.98) & 0.73 (0.73, 0.74) \\
    \bottomrule
    \end{tabular}
    \end{adjustbox}
    \end{sc}
    \end{scriptsize}
    \end{center}
    \label{tab:dataset-size}
\end{table*}

\subsection{Additional Evaluation Metrics for Online Molecular Optimization}\label{appendix:online-additional-metrics}
To characterize the generated molecules beyond
the Hit Ratio, we report validity, novelty, scaffold novelty, nearest-neighbor
similarity and the Fréchet ChemNet Distance (FCD), measured with respect to the
ZINC250K training set, for the best-performing methods on the BRAF protein target in
the online docking score optimization setting
(Section~\ref{section:online-molecular-optimization}), using 10 random seeds. We define
these metrics as follows.

\textbf{Validity.} The fraction of generated SMILES strings that RDKit~\cite{rdkit} successfully parses into a valid molecule.

\textbf{Novelty.} The fraction of generated molecules whose canonical SMILES do not appear in the training set.

\textbf{Scaffold novelty.} The fraction of generated molecules whose Bemis--Murcko scaffold~\cite{bemis1996properties} does not appear in the training set.

\textbf{Nearest-neighbor similarity.} For each generated molecule, the maximum Tanimoto similarity to the training set, computed over Morgan fingerprints~\cite{rogers2010extended} with radius 2 and 2048 bits, averaged across generated molecules.

\textbf{FCD.} The Fréchet distance between the ChemNet activation distributions of the generated and training sets~\cite{preuer2018fcd}. Lower values indicate that the generated molecules lie closer to the training distribution in chemical and bioactivity space.

All methods generate almost exclusively valid and novel molecules
(Table~\ref{tab:online-molecule-quality}). \textsc{JSI} attains scaffold novelty and
nearest-neighbor similarity comparable to SATURN, the strongest baseline in terms of
Hit Ratio, whereas GEAM generates the molecules most dissimilar from the training set.
The FCD of \textsc{JSI} is the highest, indicating that it generates molecules further
from the training distribution in chemical and bioactivity space.

\begin{table*}[h]
    \caption{Quality of the molecules generated for the BRAF protein target in the online docking score optimization setting, measured with respect to the ZINC250K training set. GEAM and NF-BO produce valid molecules by design. Values are mean (standard deviation) across 10 random seeds.}
    \begin{center}
    \begin{scriptsize}
    \begin{sc}
    \begin{adjustbox}{width=\linewidth}
    \begin{tabular}{@{}lccccc@{}}
    \toprule
    Method & Validity ($\uparrow$) & Novelty ($\uparrow$) & Scaffold Novelty ($\uparrow$) & Nearest Neighbor Similarity & FCD \\
    \midrule
    GEAM                & 1.000 (0.000) & 1.000 (0.000) & 0.990 (0.003) & 0.316 (0.007) & 16.276 (0.609) \\
    SATURN              & 0.979 (0.004) & 0.999 (0.000) & 0.931 (0.019) & 0.526 (0.020) & 27.653 (3.132) \\
    NF-BO               & 1.000 (0.000) & 0.999 (0.001) & 0.833 (0.020) & 0.502 (0.014) & 14.312 (1.589) \\
    \textsc{JSI (ours)} & 0.969 (0.008) & 0.999 (0.001) & 0.888 (0.079) & 0.542 (0.033) & 31.376 (7.209) \\
    \bottomrule
    \end{tabular}
    \end{adjustbox}
    \end{sc}
    \end{scriptsize}
    \end{center}
    \label{tab:online-molecule-quality}
\end{table*}

\subsection{Sampling Time}
\label{sec:sampling-time}

To quantify the computational cost of inference-time self-improvement, we measure
\textbf{Time}~(s), defined as the number of seconds required to sample a single solution.
Additionally, we report \textbf{\#Candidates}, the number of sequences generated per
solution, and \textbf{Length}, the mean number of generated tokens per
solution. We evaluate each sampling method on the BRAF protein target in the offline docking score optimization setting (Section~\ref{section:offline-molecular-optimization}) by sampling 100 solutions. Each method samples from the same jointly fine-tuned backbone. Unconditional sampling autoregressively generates a single sequence; Best-of-$K$ unconditionally generates $K$ sequences and returns as the solution the one with the highest prediction score, computed with the predictive part of the model; \textsc{JSI} returns a single best-scoring solution, as detailed in Algorithm~\ref{alg:joint-self-improvement}, using $N_\mathrm{rounds}$ rounds of sampling without replacement, each round producing $K$ candidates. Note that \textsc{JSI} with $N_\mathrm{rounds}=1$ is equivalent to \textsc{JSI w/o inference-time self-improvement} in Section~\ref{section:ablations}. All runs use a single NVIDIA H100 80GB GPU, and we report the
mean and standard deviation across 3 random seeds, each paired with a distinct offline
dataset.

\begin{table}[h]
    \caption{Sampling cost of the inference-time methods on the BRAF protein target in the offline docking score optimization setting. Values are mean (standard deviation) across 3 random seeds.}
    \begin{center}
    \begin{scriptsize}
    \begin{sc}
    \begin{tabular}{@{}lccc@{}}
    \toprule
    Method & \#Candidates & Length & Time(s) \\
    \midrule
    Unconditional              & 1    & 39.4 & 0.004 (0.001) \\
    Best-of-$K$                & 128  & 39.6 & 0.552 (0.014) \\
    \textsc{JSI} &&&\\
    \quad $N_\mathrm{rounds}=1$ & 128  & 39.7 & 0.662 (0.045) \\
    \quad $N_\mathrm{rounds}=2$ & 256  & 40.2 & 1.321 (0.068) \\
    \quad $N_\mathrm{rounds}=3$ & 384  & 41.0 & 2.041 (0.195) \\
    \quad $N_\mathrm{rounds}=4$ & 512  & 41.6 & 2.708 (0.154) \\
    \quad $N_\mathrm{rounds}=5$ & 640  & 42.3 & 3.418 (0.312) \\
    \quad $N_\mathrm{rounds}=6$ & 768  & 43.1 & 4.081 (0.413) \\
    \quad $N_\mathrm{rounds}=7$ & 896  & 44.0 & 4.741 (0.505) \\
    \quad $N_\mathrm{rounds}=8$ & 1024 & 44.7 & 5.440 (0.620) \\
    \bottomrule
    \end{tabular}
    \end{sc}
    \end{scriptsize}
    \end{center}
    \label{tab:sampling-time}
\end{table}

\section{Experimental Details}\label{appendix:experimental-details}

We conducted all experiments across the Helmholtz Munich and Jülich HPC clusters, using a single NVIDIA V100, A100 or H100 GPU, with 16 CPU cores, 16 GB of RAM, and a maximum wall time of 24 hours per job.

\subsection{Docking Score Optimization}

\begin{figure*}[b!]
    \centering
    \includegraphics[width=\textwidth]{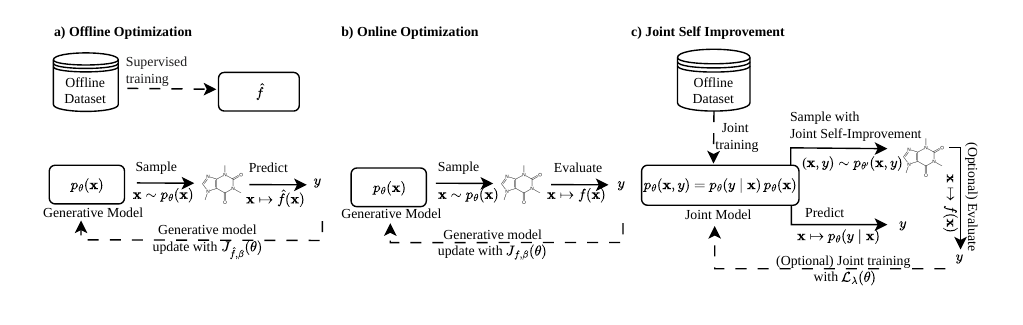}
    \vspace{-2ex}
    \caption{Offline and online GMO as compared to the \textsc{JSI} framework. (a) In the offline setting, objective $f$ is not available, and a surrogate $\hat{f}$ is used to predict its values. (b) In the online setting, objective $f$ is available and is used for generative model updates. 
    (c) \textsc{Joint Self-Improvement} can perform in both settings: either using the offline dataset to train the joint model and sample new molecules, or updating based on newly generated objective evaluations. 
}
    \label{fig:overview-experimental-setting}
    \vspace{-2ex}
\end{figure*}

To evaluate our framework under realistic sample-efficiency constraints, we adopt the experimental set-up of~\cite{lee2024drug,guo2024augmentedmemory,guo2024saturnsampleefficientgenerativemolecular,shin2025offline} with an oracle budget of 3{,}000 evaluations, since 10{,}000 evaluations~\cite{gao2022sample} may exceed the budgets typically available in real-world drug discovery workflows. We formulate docking score optimization as a constrained optimization problem, consistent with prior work~\cite{lee2024drug,guo2024augmentedmemory,guo2024saturnsampleefficientgenerativemolecular}, to generate molecules that exceed a predefined docking threshold while satisfying chemical plausibility constraints, as quantified by SA and QED. In this formulation, docking takes precedence over auxiliary properties: once the docking threshold is met, further gains in SA or QED are secondary, whereas molecules that fail to meet the threshold are rejected regardless of their auxiliary scores. Consequently, Pareto-based multi-objective optimization metrics are not well aligned with our setting, and following prior work, we use Hit Ratio, which directly measures the fraction of molecules that satisfy the docking threshold and auxiliary constraints, as our primary evaluation metric. We provide an overview of the different molecular optimization settings in 
Figure~\ref{fig:overview-experimental-setting}.

\paragraph{\textsc{JSI} Backbone}
We use Hyformer~\citep{izdebski2026synergistic} as the backbone for \textsc{JSI}. To ensure comparability with baseline methods, we pre-train Hyformer on the ZINC250K dataset, scaled to 27M parameters with $8$ transformer layers and embedding dimensionality $512$. We optimize with AdamW and a cosine learning-rate schedule with linear warmup for $20$ epochs, and select the checkpoint with the lowest validation loss. Table~\ref{table:hparams-hyformer} summarizes the pre-training hyperparameters. In contrast to~\citet{izdebski2026synergistic}, we pre-train without precomputed molecular descriptors, so that pre-training does not rely on proxies for the SA and QED feasibility constraints.

\paragraph{Docking Score Computation}
For a fair comparison with prior work, we follow the experimental setup of~\citet{lee2024drug} and use QuickVina~2~\cite{alhossary2015quickvina2} as the docking oracle in both the online and offline settings. To speed up oracle calls, we use its GPU implementation, QuickVina2-GPU~2.1. We dock against five protein targets (fa7, parp1, 5ht1b, jak2 and braf), using the same receptor structures and docking-box centers and sizes as~\citet{lee2024drug}. Each ligand is prepared from its canonical SMILES: we add hydrogens, generate a single conformer with RDKit ETKDG, minimize it with the UFF force field, and convert it to PDBQT with OpenBabel. Docking runs with the default search effort of $5000$ threads and a fixed random seed. We take the score of the lowest-energy pose as the docking score, in kcal/mol, where lower is better. Molecules that fail conformer generation are assigned a score of $0$.

\begin{table}[h]
\caption{Hyperparameter settings used for pre-training and fine-tuning of Hyformer for docking optimization task.}\label{table:hparams-hyformer}
\vskip 0.15in
\begin{center}
\begin{scriptsize}
\begin{sc}
\begin{tabular}{lcc}
\toprule
Hyperparameter & Pretraining & Finetuning \\
\midrule
Batch Size & 512 & 32  \\
Learning Rate & 6e-4 & 1e-4 \\
Weight Decay & 1e-1 & 1e-1 \\
$(\beta_1, \beta_2)$ & (0.9, 0.95) & (0.9, 0.95) \\
Gradient Clip & 1.0 & 1.0 \\
Minimum Learning Rate & 1e-6 & 1e-6 \\
Warmup Ratio & 0.05 & 0.1 \\
dtype & bfloat16 & float32 \\ 
$(p_{[\textsc{LM}]}, p_{[\textsc{MLM}]}, p_{[\textsc{PRED}]})$ & (0.9, 0.1, 0.0) & (0.75, 0.0, 0.25) \\
\bottomrule
\end{tabular}
\end{sc}
\end{scriptsize}
\end{center}
\vskip -0.1in
\end{table}

\subsection{Sample-Efficient Online Molecular Optimization}\label{appendix:online-optimization}

All results in Table~\ref{experiments:table:hit-ratio-online} are from~\cite{guo2024saturnsampleefficientgenerativemolecular}, except for NF-BO and \textsc{JSI}, which are ours.

\paragraph{NF-BO}
We run NF-BO~\cite{lee2025nfbo} ourselves, evaluating it under the same protocol applied to all other methods: the same five protein targets, the same docking oracle, and a total budget of 3,000 oracle evaluations. The generative backbone is the sequence-level autoregressive normalizing flow, with a hidden size of 500, a latent dimension of 40, 5 flow layers, 2 recurrent layers, dropout 0.2, and the non-linear squared (NLSq) transform; we initialize from the authors checkpoint pre-trained on 1.27M molecules rather than pre-training from scratch. The surrogate is a deep-kernel Gaussian process, fit for 100 epochs at initialization and 1 epoch per subsequent update, with a learning rate of 1 × $10^-3$ and a batch size of 32; the flow is updated end-to-end jointly with the surrogate for 1 epoch per update at the same learning rate. Each run is initialized from a fixed set of 100 pre-scored molecules, so that 2,900 of the 3,000 evaluations are spent on molecules proposed during optimization, and an end-to-end update of the flow and surrogate is triggered approximately every 200 oracle evaluations. Candidates are acquired by Thompson sampling from the 5 highest-scoring molecules in the current dataset, drawing a batch of 10 candidates from each and capping the candidate pool at 2,000, which yields 50 molecules scored by the oracle per round. We use a sampling temperature of 0.01 and a DACS temperature of 400. Each target is repeated over 10 random seeds, and we report the mean and standard deviation across runs.

\paragraph{\textsc{JSI}} We use Hyformer as the generative model, pre-trained on ZINC250K, to sample molecules using inference-time self-improvement. Oracle queries are employed to compute the reward. The generator is then fine-tuned using the top rewarded molecules. Following~\cite{guo2024saturnsampleefficientgenerativemolecular}, 5 rounds of SMILES augmentation is applied prior to fine-tuning. We sample molecules at a temperature of $0.9$. We perform a grid search over the beam width $K \in \{16, 32, 64, 128\}$, the number of rounds $N_{\mathrm{rounds}} \in \{4, 8, 10, 12, 14\}$, and the step size $\sigma \in \{0.25, 0.5, 0.75, 1.0, 1.25, 1.5, 1.75\}$. Best hyperparameters across all protein targets are summarized in Table~\ref{table:joint-self-improvement-hparams-online}.

\begin{table*}[!t]
\centering
\caption{Best hyperparameters for Joint Self-Improvement in the online optimization experiment.}
\label{table:joint-self-improvement-hparams-online}
\begin{scriptsize}
\begin{tabular}{lccccc}
\toprule
& \multicolumn{5}{c}{Target Protein} \\
\cmidrule{2-6}
Hyperparameter & parp1 & fa7 & 5ht1b & braf & jak2 \\
\midrule
Beam width $K$ & 16 & 16 & 32 & 16 & 16 \\
Number of rounds $N_\textrm{rounds}$ & 10 & 14 & 10 & 12 & 10 \\
Step size $\sigma$ & 1.5 & 1.5 & 0.75 & 1.75 & 1.5  \\
\bottomrule
\end{tabular}
\end{scriptsize}
\end{table*}

\subsection{Sample-Efficient Offline Molecular Optimization}\label{appendix:offline-optimization}

All results in Table~\ref{table:offline-optimization} are ours. In Figure~\ref{fig:docked_poses}, docking poses are visualized using PyMOL.

\paragraph{GraphGA} We use the same Hyformer backbone as JSI for candidate generation and surrogate scoring, replacing inference-time self-improvement with graph-based genetic search. At each optimization round, we initialize the population with the highest-scoring generated molecules and apply rank-based parent selection, crossover, and mutation with probability 0.01.

\paragraph{REINVENT}

We implement a surrogate-guided REINVENT optimization pipeline with an integrated Hyformer predictor. First Hyformer is fine-tuned on the offline training split, and the resulting predictor is used to score generated molecules throughout the optimization. We initialize a REINVENT agent from a fixed prior checkpoint pre-trained on the  ZINC250K dataset and then run iterative policy updates: at each step, the agent samples a batch of generated molecules and Hyformer predicts the reward. The agent is then updated by minimizing the REINVENT squared error between the agent's log-likelihood and an augmented log-likelihood target, with experience replay. All the hyperparameters for the experiment are taken from \cite{shin2025offline}.

\paragraph{SATURN}
We implement a surrogate-guided SATURN optimization pipeline with an integrated Hyformer predictor, trained using a REINVENT-style reinforcement learning objective with augmented memory replay and SMILES-based data augmentation. Hyformer is first fine-tuned on the offline training dataset and then used as a surrogate oracle during optimization: at each iteration, molecules are sampled from the current policy, scored by Hyformer, and the policy is updated based on the predicted rewards. The SATURN agent is initialized from a prior pre-trained on ZINC250K, using a publicly available checkpoint from the SATURN repository~\url{https://github.com/schwallergroup/saturn}. Following~\cite{guo2024saturnsampleefficientgenerativemolecular}, we use a learning rate of $1\times10^{-4}$, an augmented-likelihood scaling of $128.0$ (denoted $\sigma$ by~\citet{guo2024saturnsampleefficientgenerativemolecular}, and distinct from the \textsc{JSI} step size $\sigma$), 10 SMILES augmentation rounds, a replay buffer of size 100, and a batch size of 64.

\paragraph{GeneticGFN}
We first fine-tune Hyformer on the offline dataset using hyperparameters summarized in Table~\ref{table:hparams-hyformer} and employ the resulting model as a surrogate predictor throughout optimization. Following~\citet{shin2025offline}, we perform 8 augmentation rounds. In each round, we sample 200 molecules from the generative policy and score them using the surrogate. Across all rounds, the top 64 highest-scoring molecules are used to initialize a genetic algorithm, which is run for 2 generations with a mutation rate of 0.01 and an offspring size of 4. All generated offspring are evaluated by the surrogate and added to a rank-based reweighted experience buffer with a maximum capacity of 512. At the end of each augmentation round the policy is optimized using the trajectory balance loss with a Kullback-Leibler (KL) divergence penalty. We use a KL coefficient of $1\times10^{-2}$ and a learning rate of $5\times10^{-4}$ for both the policy parameters and the log-partition constant logZ, which is initialized to $1\times10^{-3}$.

\paragraph{RaM}
We fine-tune Hyformer on the offline dataset to serve as a learning-to-rank (LTR) surrogate model using the RankCosine loss. We use a batch size of 32 and a learning rate of $1\times10^{-4}$, with all other hyperparameters set to their default values (Table~\ref{table:hparams-hyformer}). The best-scoring sequences in the offline dataset, i.e., those with the lowest docking scores, are selected as initial candidates for gradient-ascent optimization in RaM. The optimization is performed using stochastic gradient ascent for 500 steps with a step size of 0.8, where the target score is defined as the mean of the normalized LTR objectives, and gradients are applied at the token-embedding level. The optimized embeddings are mapped back to discrete tokens using cosine similarity. The step size and initial candidate selection strategy are determined via hyperparameter optimization. At each iteration, generated sequences are aggregated and filtered to retain only valid and unique sequences that are also not present in the offline dataset. The top-ranked sequences according to the LTR surrogate are then selected for final evaluation.

\paragraph{\textsc{JSI}}
We jointly fine-tune Hyformer on the offline dataset with the hyperparameters listed in Table~\ref{table:hparams-hyformer}. Crucially, we fine-tune on Z-score normalized property values, with the mean and standard deviation of each property computed from the offline dataset, so that the prediction head directly outputs normalized values. Next, we define the objective as the mean of the predicted normalized properties (Eq.~\eqref{eq:reward-offline}), where DS and SA enter with a negative sign since lower values are better. The advantage of a sampled molecule is its reward minus the expected objective over the current beam, min-max normalized to $[-1, 1]$ separately for positive and negative advantages, so that the expected objective maps to $0$. Finally, we sample the required number of molecules for evaluation with temperature $0.9$  and perform a grid search over beam width $K \in \{16, 32, 64, 128\}$, number of rounds $N_\textrm{rounds} \in \{4, 8, 10, 12\}$ and step size $\sigma \in \{0.25, 0.5, 0.75, 1.0, 1.25, 1.5\}$. The best hyperparameters are summarized in Table~\ref{table:joint-self-improvement-hparams-offline}. We note that the best hyperparameter settings are fairly consistent across different protein targets, indicating their transferability across tasks in the offline setting. To ensure numerical correctness and efficiency of our sampling algorithm, in particular numerical stability in Eq.~\eqref{eq:perturbed-model-practical}, we follow the augmented trie implementation of~\citet{shi2021incrementalsamplingreplacementsequence}.

\begin{table*}[!t]
\centering
\caption{Best hyperparameters for Joint Self-Improvement in the offline optimization experiment.}
\label{table:joint-self-improvement-hparams-offline}
\begin{scriptsize}
\begin{tabular}{lccccc}
\toprule
& \multicolumn{5}{c}{Target Protein} \\
\cmidrule{2-6}
Hyperparameter & parp1 & fa7 & 5ht1b & braf & jak2 \\
\midrule
Beam width $K$ & 128 & 128 & 16 & 128 & 128 \\
Number of rounds $N_\textrm{rounds}$ & 10 & 10 & 8 & 8 & 8\\
Step size $\sigma$ & 0.25 & 0.5 & 1.0 & 1.0 & 0.5 \\
\bottomrule
\end{tabular}
\end{scriptsize}
\end{table*}

\subsection{Sample-Efficient Offline AMP Optimization}\label{appendix:amp}

All results in Table~\ref{tab:amp_results} and Table~\ref{tab:amp_results_harder} are ours. For all baseline methods, except for \textsc{JSI} and Hyformer, we randomly subsample 1{,}000 sequences from the full set of generated peptides (50{,}000 sequences per method) provided by~\cite{chen2023ampdiffusion}, in order to match a fixed evaluation budget of 1{,}000 sequences across methods.

\paragraph{Hyformer}
Peptide sequences are generated using rejection sampling following~\cite{izdebski2026synergistic}. Specifically, sequences are sampled from the generative component and accepted using the predictive component with a threshold of $\mathrm{MIC} \leq 10^{2.0}$ until 1{,}000 sequences are collected. No additional retraining is performed during generation.

\paragraph{JSI}
We use Hyformer~\cite{izdebski2026synergistic} as the backbone model for all experiments. The model is pre-trained on 3.5M peptide sequences and subsequently on an additional 1M AMP sequences, using the hyperparameters listed in Table~\ref{table:hparams-amp}. The offline dataset consists of 3{,}444 peptides with associated MIC values against \textit{E.~coli}, following~\cite{Szymczak2023HydrAMP}. We jointly fine-tune the model on the offline dataset using the joint loss $\mathcal{L}_\lambda$ with $\lambda = 0.4$ (Equation~\eqref{eq:joint-loss}).

We perform inference-time self-improvement by biasing the generative component using the predictive component as a reward, defined as $R(x) = -\hat{\mathrm{MIC}}(x)$. We iteratively sample peptide sequences using \textsc{JSI} until 1{,}000 sequences are collected, without further retraining. We use a beam width of 32, the number of self-improvement rounds $N_{\mathrm{rounds}} = 8$, step size $\sigma = 1.0$ and with temperature 0.9. 

For Table~\ref{tab:amp_results_harder}, we additionally evaluate a more challenging setting where the offline dataset is reconstructed based on activity labels defined in prior work~\cite{Szymczak2023HydrAMP}, resulting in datasets containing only 1\% and 0\% active peptides. In both settings, we fine-tune the Hyformer backbone using the same hyperparameters listed in Table~\ref{table:hparams-amp}. Hyformer uses rejection sampling with the same MIC threshold, while \textsc{JSI} performs inference-time self-improvement without retraining.

We estimate uncertainty using bootstrapping with 1{,}000 resampling iterations with replacement over the selected 1{,}000 sequences. For each metric, we report the mean computed on the full set of sequences and construct 95\% confidence intervals using the 2.5th and 97.5th percentiles of the bootstrap distribution. For the diversity metric, we apply the same procedure by resampling sequences and recomputing the diversity score for each bootstrap sample.

\begin{table}[h]
\caption{Hyperparameter settings used for pre-training and fine-tuning of Hyformer for AMP optimization task.}\label{table:hparams-amp}
\vskip 0.15in
\begin{center}
\begin{scriptsize}
\begin{sc}
\begin{tabular}{lcc}
\toprule
Hyperparameter & Pretraining & Finetuning \\
\midrule
Batch Size & 512 & 64  \\
Learning Rate & 4e-4 & 1e-4 \\
Weight Decay & 1e-1 & 1e-1 \\
$(\beta_1, \beta_2)$ & (0.9, 0.95) & (0.9, 0.95) \\
Gradient Clip & 1.0 & 1.0 \\
Minimum Learning Rate & 1e-6 & 1e-6 \\
Warmup Ratio & 0.05 & 0.1 \\
dtype & bfloat16 & float32 \\ 
$(p_{[\textsc{LM}]}, p_{[\textsc{MLM}]}, p_{[\textsc{PRED}]})$ & (0.9, 0.1, 0.0) & (0.6, 0.0, 0.4) \\
\bottomrule
\end{tabular}
\end{sc}
\end{scriptsize}
\end{center}
\vskip -0.1in
\end{table}

\subsubsection{Benchmark metrics}\label{appendix:amp-metrics}
Following~\cite{izdebski2026synergistic}, we use six metrics to evaluate the generated AMPs.

\paragraph{Perplexity} Exponentiated token-level negative log-likelihood of a sequence, computed using ProGen2-medium~\cite{nijkamp2023progen2}, following~\citet{torres2025ampdiffusion}; lower values indicate higher plausibility of the generated peptides under the model.
\paragraph{Diversity} The average pairwise Levenshtein distance among generated sequences, where higher values reflect increased diversity. We follow Eq.~6 in~\cite{kim2021drugdiscovery}, replacing the Soergel distance with Levenshtein distance in the Hyformer setting.
\paragraph{Fitness} A quantitative measure of a peptide's tendency to form a stable, amphipathic $\alpha$-helical structure, as defined in~\cite{li2024foundationamp}.
\paragraph{HydrAMP$_{\text{MIC}}$}The predicted probability that a peptide is active against \textit{E.~coli}, computed using HydrAMP~\cite{Szymczak2023HydrAMP}.
\paragraph{AMPlify} The predicted probability that a peptide is an AMP, computed using AMPlify~\cite{li2022amplify}.
\paragraph{AmPEPpy} The predicted probability that a peptide is an AMP, computed using AmPEPpy~\cite{lawrence2020ampeppy}.

\subsection{Ablations}\label{appendix:ablations}

All results in Table~\ref{table:ablations} are ours.

\paragraph{\textsc{JSI w/o Joint Modeling}}
We train Chemprop~\cite{heid2023chemprop} as the surrogate predictor on the 
offline dataset with default hyperparameters~\cite{heid2023chemprop}, including a batch size of 64, a 
maximum learning rate of $10^{-3}$, and up to 30 epochs with early stopping. 
For sampling, we fine-tune the pre-trained Hyformer on the offline dataset using the
generative-only part of the joint loss, $\mathcal{L}_{\lambda = 0}$ (Eq.~\eqref{eq:joint-loss}), so that
the comparison isolates the effect of joint modeling, with the best
hyperparameters from Appendix~\ref{appendix:online-optimization}. To verify that Chemprop provides a competitive surrogate, we evaluate predictive performance
on a held-out test set of 500 molecules randomly sampled from ZINC250K
(Table~\ref{table:surrogates}).

\paragraph{\textsc{JSI w/o inference-time self-improvement}} We replace the joint self-improving sampling scheme with Best-of-$K$ sampling~\citep{izdebski2026synergistic}, where for every molecule we sample $K$ candidates using the generative part of the model and select the one with the best predicted reward. Crucially, we draw the $K$ candidates efficiently in a single round of stochastic beam search, which yields $K$ distinct sequences without replacement while sharing computation across common prefixes, and we apply no advantage-based biasing of the generative model, i.e. $N_{\textrm{rounds}} = 1$. Predictor-based selection of the best candidate, which \textsc{JSI} also performs (Step~\ref{step:selection} in Algorithm~\ref{alg:joint-self-improvement}), is retained, so that the comparison isolates the effect of inference-time self-improvement. We keep the beam width $K$ selected for \textsc{JSI} on each target, i.e. $K = 128$ for parp1, fa7, braf and jak2, and $K = 16$ for 5HT1B, and leave all remaining hyperparameters unchanged.

\begin{table*}[h]
    \caption{Predictive performance of Chemprop and  \textsc{JSI} 
on a held-out test set of 500 molecules sampled from ZINC250K in docking score 
prediction, as measured by mean absolute error (MAE).}
    \begin{center}
    \begin{scriptsize}
    \begin{sc}
    \begin{tabular}{@{}lccccc@{}}
    \toprule
    & \multicolumn{5}{c}{Target Protein, MAE ($\downarrow$)} \\
    \cmidrule{2-6}
    Surrogate & parp1 & fa7 & 5ht1b & braf & jak2 \\
    \midrule
    Chemprop & 0.517 (0.012) & 0.383 (0.005) & 0.528 (0.007) & 0.464 (0.006) & 0.439 (0.013) \\
    \textsc{JSI} & 0.446 (0.013) & 0.400 (0.012) & 0.426 (0.015) & 0.466 (0.012) & 0.408 (0.007) \\
    \bottomrule
    \end{tabular}
    \end{sc}
    \end{scriptsize}
    \end{center}
    \label{table:surrogates}
\end{table*}

\newpage
\input{checklist.tex}

\end{document}

%% file: checklist.tex
\section*{NeurIPS Paper Checklist}

\begin{enumerate}

\item {\bf Claims}
    \item[] Question: Do the main claims made in the abstract and introduction accurately reflect the paper's contributions and scope?
    \item[] Answer: \answerYes{} %
    \item[] Justification: We evaluate all the claims via experimental evaluation in Section~\ref{section:experiments} and in Appendix~\ref{appendix:REINVENT-gradient}
    \item[] Guidelines:
    \begin{itemize}
        \item The answer \answerNA{} means that the abstract and introduction do not include the claims made in the paper.
        \item The abstract and/or introduction should clearly state the claims made, including the contributions made in the paper and important assumptions and limitations. A \answerNo{} or \answerNA{} answer to this question will not be perceived well by the reviewers. 
        \item The claims made should match theoretical and experimental results, and reflect how much the results can be expected to generalize to other settings. 
        \item It is fine to include aspirational goals as motivation as long as it is clear that these goals are not attained by the paper. 
    \end{itemize}

\item {\bf Limitations}
    \item[] Question: Does the paper discuss the limitations of the work performed by the authors?
    \item[] Answer: \answerYes{} %
    \item[] Justification: We discuss limitations in Section~\ref{section:conclusions}. Moreover, we discuss the running time of our algorithm in paragraph \textbf{Practical Implementation} and in ablations. 
    \item[] Guidelines:
    \begin{itemize}
        \item The answer \answerNA{} means that the paper has no limitation while the answer \answerNo{} means that the paper has limitations, but those are not discussed in the paper. 
        \item The authors are encouraged to create a separate ``Limitations'' section in their paper.
        \item The paper should point out any strong assumptions and how robust the results are to violations of these assumptions (e.g., independence assumptions, noiseless settings, model well-specification, asymptotic approximations only holding locally). The authors should reflect on how these assumptions might be violated in practice and what the implications would be.
        \item The authors should reflect on the scope of the claims made, e.g., if the approach was only tested on a few datasets or with a few runs. In general, empirical results often depend on implicit assumptions, which should be articulated.
        \item The authors should reflect on the factors that influence the performance of the approach. For example, a facial recognition algorithm may perform poorly when image resolution is low or images are taken in low lighting. Or a speech-to-text system might not be used reliably to provide closed captions for online lectures because it fails to handle technical jargon.
        \item The authors should discuss the computational efficiency of the proposed algorithms and how they scale with dataset size.
        \item If applicable, the authors should discuss possible limitations of their approach to address problems of privacy and fairness.
        \item While the authors might fear that complete honesty about limitations might be used by reviewers as grounds for rejection, a worse outcome might be that reviewers discover limitations that aren't acknowledged in the paper. The authors should use their best judgment and recognize that individual actions in favor of transparency play an important role in developing norms that preserve the integrity of the community. Reviewers will be specifically instructed to not penalize honesty concerning limitations.
    \end{itemize}

\item {\bf Theory assumptions and proofs}
    \item[] Question: For each theoretical result, does the paper provide the full set of assumptions and a complete (and correct) proof?
    \item[] Answer: \answerNA{}{} %
    \item[] Justification: The paper does not contain novel theoretical results.
    \item[] Guidelines:
    \begin{itemize}
        \item The answer \answerNA{} means that the paper does not include theoretical results. 
        \item All the theorems, formulas, and proofs in the paper should be numbered and cross-referenced.
        \item All assumptions should be clearly stated or referenced in the statement of any theorems.
        \item The proofs can either appear in the main paper or the supplemental material, but if they appear in the supplemental material, the authors are encouraged to provide a short proof sketch to provide intuition. 
        \item Inversely, any informal proof provided in the core of the paper should be complemented by formal proofs provided in appendix or supplemental material.
        \item Theorems and Lemmas that the proof relies upon should be properly referenced. 
    \end{itemize}

    \item {\bf Experimental result reproducibility}
    \item[] Question: Does the paper fully disclose all the information needed to reproduce the main experimental results of the paper to the extent that it affects the main claims and/or conclusions of the paper (regardless of whether the code and data are provided or not)?
    \item[] Answer: \answerYes{} %
    \item[] Justification: We provide all necessary details for reproducing all the results in Appendix~\ref{appendix:experimental-details}.
    \item[] Guidelines:
    \begin{itemize}
        \item The answer \answerNA{} means that the paper does not include experiments.
        \item If the paper includes experiments, a \answerNo{} answer to this question will not be perceived well by the reviewers: Making the paper reproducible is important, regardless of whether the code and data are provided or not.
        \item If the contribution is a dataset and\slash or model, the authors should describe the steps taken to make their results reproducible or verifiable. 
        \item Depending on the contribution, reproducibility can be accomplished in various ways. For example, if the contribution is a novel architecture, describing the architecture fully might suffice, or if the contribution is a specific model and empirical evaluation, it may be necessary to either make it possible for others to replicate the model with the same dataset, or provide access to the model. In general. releasing code and data is often one good way to accomplish this, but reproducibility can also be provided via detailed instructions for how to replicate the results, access to a hosted model (e.g., in the case of a large language model), releasing of a model checkpoint, or other means that are appropriate to the research performed.
        \item While NeurIPS does not require releasing code, the conference does require all submissions to provide some reasonable avenue for reproducibility, which may depend on the nature of the contribution. For example
        \begin{enumerate}
            \item If the contribution is primarily a new algorithm, the paper should make it clear how to reproduce that algorithm.
            \item If the contribution is primarily a new model architecture, the paper should describe the architecture clearly and fully.
            \item If the contribution is a new model (e.g., a large language model), then there should either be a way to access this model for reproducing the results or a way to reproduce the model (e.g., with an open-source dataset or instructions for how to construct the dataset).
            \item We recognize that reproducibility may be tricky in some cases, in which case authors are welcome to describe the particular way they provide for reproducibility. In the case of closed-source models, it may be that access to the model is limited in some way (e.g., to registered users), but it should be possible for other researchers to have some path to reproducing or verifying the results.
        \end{enumerate}
    \end{itemize}

\item {\bf Open access to data and code}
    \item[] Question: Does the paper provide open access to the data and code, with sufficient instructions to faithfully reproduce the main experimental results, as described in supplemental material?
    \item[] Answer: \answerYes{} %
    \item[] Justification: The repository with scripts reproducing the results is hosted on GitHub and will be made publicly available upon acceptance. 
    \item[] Guidelines:
    \begin{itemize}
        \item The answer \answerNA{} means that paper does not include experiments requiring code.
        \item Please see the NeurIPS code and data submission guidelines (\url{https://neurips.cc/public/guides/CodeSubmissionPolicy}) for more details.
        \item While we encourage the release of code and data, we understand that this might not be possible, so \answerNo{} is an acceptable answer. Papers cannot be rejected simply for not including code, unless this is central to the contribution (e.g., for a new open-source benchmark).
        \item The instructions should contain the exact command and environment needed to run to reproduce the results. See the NeurIPS code and data submission guidelines (\url{https://neurips.cc/public/guides/CodeSubmissionPolicy}) for more details.
        \item The authors should provide instructions on data access and preparation, including how to access the raw data, preprocessed data, intermediate data, and generated data, etc.
        \item The authors should provide scripts to reproduce all experimental results for the new proposed method and baselines. If only a subset of experiments are reproducible, they should state which ones are omitted from the script and why.
        \item At submission time, to preserve anonymity, the authors should release anonymized versions (if applicable).
        \item Providing as much information as possible in supplemental material (appended to the paper) is recommended, but including URLs to data and code is permitted.
    \end{itemize}

\item {\bf Experimental setting/details}
    \item[] Question: Does the paper specify all the training and test details (e.g., data splits, hyperparameters, how they were chosen, type of optimizer) necessary to understand the results?
    \item[] Answer: \answerYes{} %
    \item[] Justification: We provide all necessary details for reproducing all the results in Appendix~\ref{appendix:experimental-details}.
    \item[] Guidelines:
    \begin{itemize}
        \item The answer \answerNA{} means that the paper does not include experiments.
        \item The experimental setting should be presented in the core of the paper to a level of detail that is necessary to appreciate the results and make sense of them.
        \item The full details can be provided either with the code, in appendix, or as supplemental material.
    \end{itemize}

\item {\bf Experiment statistical significance}
    \item[] Question: Does the paper report error bars suitably and correctly defined or other appropriate information about the statistical significance of the experiments?
    \item[] Answer: \answerYes{} %
    \item[] Justification: Across all experiments, we report mean and standard deviation over 10 random seeds, or bootstrapped confidence intervals.
    \item[] Guidelines:
    \begin{itemize}
        \item The answer \answerNA{} means that the paper does not include experiments.
        \item The authors should answer \answerYes{} if the results are accompanied by error bars, confidence intervals, or statistical significance tests, at least for the experiments that support the main claims of the paper.
        \item The factors of variability that the error bars are capturing should be clearly stated (for example, train/test split, initialization, random drawing of some parameter, or overall run with given experimental conditions).
        \item The method for calculating the error bars should be explained (closed form formula, call to a library function, bootstrap, etc.)
        \item The assumptions made should be given (e.g., Normally distributed errors).
        \item It should be clear whether the error bar is the standard deviation or the standard error of the mean.
        \item It is OK to report 1-sigma error bars, but one should state it. The authors should preferably report a 2-sigma error bar than state that they have a 96\% CI, if the hypothesis of Normality of errors is not verified.
        \item For asymmetric distributions, the authors should be careful not to show in tables or figures symmetric error bars that would yield results that are out of range (e.g., negative error rates).
        \item If error bars are reported in tables or plots, the authors should explain in the text how they were calculated and reference the corresponding figures or tables in the text.
    \end{itemize}

\item {\bf Experiments compute resources}
    \item[] Question: For each experiment, does the paper provide sufficient information on the computer resources (type of compute workers, memory, time of execution) needed to reproduce the experiments?
    \item[] Answer: \answerYes{} %
    \item[] Justification: We provide all necessary details for reproducing all the results in Appendix~\ref{appendix:experimental-details}.
    \item[] Guidelines:
    \begin{itemize}
        \item The answer \answerNA{} means that the paper does not include experiments.
        \item The paper should indicate the type of compute workers CPU or GPU, internal cluster, or cloud provider, including relevant memory and storage.
        \item The paper should provide the amount of compute required for each of the individual experimental runs as well as estimate the total compute. 
        \item The paper should disclose whether the full research project required more compute than the experiments reported in the paper (e.g., preliminary or failed experiments that didn't make it into the paper). 
    \end{itemize}
    
\item {\bf Code of ethics}
    \item[] Question: Does the research conducted in the paper conform, in every respect, with the NeurIPS Code of Ethics \url{https://neurips.cc/public/EthicsGuidelines}?
    \item[] Answer: \answerYes{} %
    \item[] Justification: The research conducted in the paper conforms with the NeurIPS Code of Ethics.
    \item[] Guidelines:
    \begin{itemize}
        \item The answer \answerNA{} means that the authors have not reviewed the NeurIPS Code of Ethics.
        \item If the authors answer \answerNo, they should explain the special circumstances that require a deviation from the Code of Ethics.
        \item The authors should make sure to preserve anonymity (e.g., if there is a special consideration due to laws or regulations in their jurisdiction).
    \end{itemize}

\item {\bf Broader impacts}
    \item[] Question: Does the paper discuss both potential positive societal impacts and negative societal impacts of the work performed?
    \item[] Answer: \answerYes{} %
    \item[] Justification: We address the potential negative societal impacts. 
    \item[] Guidelines:
    \begin{itemize}
        \item The answer \answerNA{} means that there is no societal impact of the work performed.
        \item If the authors answer \answerNA{} or \answerNo, they should explain why their work has no societal impact or why the paper does not address societal impact.
        \item Examples of negative societal impacts include potential malicious or unintended uses (e.g., disinformation, generating fake profiles, surveillance), fairness considerations (e.g., deployment of technologies that could make decisions that unfairly impact specific groups), privacy considerations, and security considerations.
        \item The conference expects that many papers will be foundational research and not tied to particular applications, let alone deployments. However, if there is a direct path to any negative applications, the authors should point it out. For example, it is legitimate to point out that an improvement in the quality of generative models could be used to generate Deepfakes for disinformation. On the other hand, it is not needed to point out that a generic algorithm for optimizing neural networks could enable people to train models that generate Deepfakes faster.
        \item The authors should consider possible harms that could arise when the technology is being used as intended and functioning correctly, harms that could arise when the technology is being used as intended but gives incorrect results, and harms following from (intentional or unintentional) misuse of the technology.
        \item If there are negative societal impacts, the authors could also discuss possible mitigation strategies (e.g., gated release of models, providing defenses in addition to attacks, mechanisms for monitoring misuse, mechanisms to monitor how a system learns from feedback over time, improving the efficiency and accessibility of ML).
    \end{itemize}
    
\item {\bf Safeguards}
    \item[] Question: Does the paper describe safeguards that have been put in place for responsible release of data or models that have a high risk for misuse (e.g., pre-trained language models, image generators, or scraped datasets)?
    \item[] Answer: \answerNA{} %
    \item[] Justification: The paper does not release new pretrained checkpoints, nor data.
    \item[] Guidelines:
    \begin{itemize}
        \item The answer \answerNA{} means that the paper poses no such risks.
        \item Released models that have a high risk for misuse or dual-use should be released with necessary safeguards to allow for controlled use of the model, for example by requiring that users adhere to usage guidelines or restrictions to access the model or implementing safety filters. 
        \item Datasets that have been scraped from the Internet could pose safety risks. The authors should describe how they avoided releasing unsafe images.
        \item We recognize that providing effective safeguards is challenging, and many papers do not require this, but we encourage authors to take this into account and make a best faith effort.
    \end{itemize}

\item {\bf Licenses for existing assets}
    \item[] Question: Are the creators or original owners of assets (e.g., code, data, models), used in the paper, properly credited and are the license and terms of use explicitly mentioned and properly respected?
    \item[] Answer: \answerNA{} %
    \item[] Justification: No protected assets were used. 
    \item[] Guidelines:
    \begin{itemize}
        \item The answer \answerNA{} means that the paper does not use existing assets.
        \item The authors should cite the original paper that produced the code package or dataset.
        \item The authors should state which version of the asset is used and, if possible, include a URL.
        \item The name of the license (e.g., CC-BY 4.0) should be included for each asset.
        \item For scraped data from a particular source (e.g., website), the copyright and terms of service of that source should be provided.
        \item If assets are released, the license, copyright information, and terms of use in the package should be provided. For popular datasets, \url{paperswithcode.com/datasets} has curated licenses for some datasets. Their licensing guide can help determine the license of a dataset.
        \item For existing datasets that are re-packaged, both the original license and the license of the derived asset (if it has changed) should be provided.
        \item If this information is not available online, the authors are encouraged to reach out to the asset's creators.
    \end{itemize}

\item {\bf New assets}
    \item[] Question: Are new assets introduced in the paper well documented and is the documentation provided alongside the assets?
    \item[] Answer: \answerNA{} %
    \item[] Justification: The paper does not release new assets.
    \item[] Guidelines:
    \begin{itemize}
        \item The answer \answerNA{} means that the paper does not release new assets.
        \item Researchers should communicate the details of the dataset\slash code\slash model as part of their submissions via structured templates. This includes details about training, license, limitations, etc. 
        \item The paper should discuss whether and how consent was obtained from people whose asset is used.
        \item At submission time, remember to anonymize your assets (if applicable). You can either create an anonymized URL or include an anonymized zip file.
    \end{itemize}

\item {\bf Crowdsourcing and research with human subjects}
    \item[] Question: For crowdsourcing experiments and research with human subjects, does the paper include the full text of instructions given to participants and screenshots, if applicable, as well as details about compensation (if any)? 
    \item[] Answer: \answerNA{} %
    \item[] Justification: The paper does not involve crowdsourcing nor research with human subjects.
    \item[] Guidelines:
    \begin{itemize}
        \item The answer \answerNA{} means that the paper does not involve crowdsourcing nor research with human subjects.
        \item Including this information in the supplemental material is fine, but if the main contribution of the paper involves human subjects, then as much detail as possible should be included in the main paper. 
        \item According to the NeurIPS Code of Ethics, workers involved in data collection, curation, or other labor should be paid at least the minimum wage in the country of the data collector. 
    \end{itemize}

\item {\bf Institutional review board (IRB) approvals or equivalent for research with human subjects}
    \item[] Question: Does the paper describe potential risks incurred by study participants, whether such risks were disclosed to the subjects, and whether Institutional Review Board (IRB) approvals (or an equivalent approval/review based on the requirements of your country or institution) were obtained?
    \item[] Answer: \answerNA{} %
    \item[] Justification: The paper does not involve crowdsourcing nor research with human subjects.
    \item[] Guidelines:
    \begin{itemize}
        \item The answer \answerNA{} means that the paper does not involve crowdsourcing nor research with human subjects.
        \item Depending on the country in which research is conducted, IRB approval (or equivalent) may be required for any human subjects research. If you obtained IRB approval, you should clearly state this in the paper. 
        \item We recognize that the procedures for this may vary significantly between institutions and locations, and we expect authors to adhere to the NeurIPS Code of Ethics and the guidelines for their institution. 
        \item For initial submissions, do not include any information that would break anonymity (if applicable), such as the institution conducting the review.
    \end{itemize}

\item {\bf Declaration of LLM usage}
    \item[] Question: Does the paper describe the usage of LLMs if it is an important, original, or non-standard component of the core methods in this research? Note that if the LLM is used only for writing, editing, or formatting purposes and does \emph{not} impact the core methodology, scientific rigor, or originality of the research, declaration is not required.
    \item[] Answer: \answerNA{} %
    \item[] Justification: LLMs were used only for writing, editing, or formatting purposes.
    \item[] Guidelines:
    \begin{itemize}
        \item The answer \answerNA{} means that the core method development in this research does not involve LLMs as any important, original, or non-standard components.
        \item Please refer to our LLM policy in the NeurIPS handbook for what should or should not be described.
    \end{itemize}

\end{enumerate}